%% file: main.tex
\documentclass[a4paper, 10pt, conference]{ieeeconf}      
\usepackage{FG2021}
\usepackage{graphicx}
\usepackage{hyperref}
\usepackage{multirow}
\usepackage{subcaption}
\usepackage[accsupp]{axessibility} 
\FGfinalcopy 

\IEEEoverridecommandlockouts                              
\def\FGPaperID{265} 

\title{\LARGE \bf A Study of the Human Perception of Synthetic Faces}

\author{\parbox{16cm}{\centering
    {\large Bingyu Shen$^1$, Brandon RichardWebster$^2$, Alice O'Toole$^3$, Kevin Bowyer$^1$ \\
    and Walter J. Scheirer$^1$}\\
    {\normalsize
    $^1$ Department of Computer Science and Engineering, University of Notre Dame, Notre Dame, IN, USA\\
    $^2$ Kitware, Inc., Clifton Park, NY, USA\\
    $^3$ School of Behavioral and Brain Sciences, The University of Texas at Dallas, Richardson, TX, USA
    }}
}

\begin{document}

\IEEEoverridecommandlockouts\pubid{\makebox[\columnwidth]{978-1-6654-3176-7/21/\$31.00~\copyright{}2021 IEEE \hfill} \hspace{\columnsep}\makebox[\columnwidth]{ }}

\ifFGfinal
\thispagestyle{empty}
\pagestyle{empty}
\else
\author{Anonymous FG2021 submission\\ Paper ID \FGPaperID \\}
\pagestyle{plain}
\fi
\maketitle

\input{1abstract}
\input{2introduction}
\input{3background}

\input{4method}
\input{5experiment}

\input{6conclusions}

\input{7acknowledgments}

{\small
\bibliographystyle{ieee}
\bibliography{egbib}
}

\end{document}

%% file: 1abstract.tex
\begin{abstract}

Advances in face synthesis have raised alarms about the deceptive use of synthetic faces. Can synthetic identities be effectively used to fool human observers?
In this paper, we introduce a study of the human perception of synthetic faces generated using different strategies including a state-of-the-art deep learning-based GAN model. This is the first rigorous study of the effectiveness of synthetic face generation techniques grounded in experimental techniques from psychology. We answer important questions such as how often do GAN-based and more traditional image processing-based techniques confuse human observers, and are there subtle cues within a synthetic face image that cause humans to perceive it as a fake without having to search for obvious clues? To answer these questions, we conducted a series of large-scale crowdsourced behavioral experiments with different sources of face imagery. Results show that humans are unable to distinguish synthetic faces from real faces under several different circumstances. This finding has serious implications for many different applications where face images are presented to human users. 

\end{abstract}

%% file: 2introduction.tex
\section{Introduction}

Face synthesis is a popular computer vision approach with many creative applications, but a few potentially dangerous ones as well. The purpose of face synthesis is to generate photo-realistic face images of non-existent identities from real face input, and several generative methods have achieved great success in this endeavour~\cite{he2019attgan, karras2019style, liu2019stgan}. This family of techniques includes, but is not limited to, face swapping~\cite{bitouk2008face, nirkin2018face, nirkin2019fsgan}, face aging~\cite{genovese2019towards, liu2017face}, and face expression synthesis~\cite{zhang2005geometry}. Recently, the resolution and quality of synthetic human faces has drastically improved thanks to the development of generative adversarial networks~(GANs)~\cite{karras2019style, song2018geometry, zhao2017dual}. Remarkably, the plausibility of generated faces from these techniques has not been assessed in any rigorous way. Is it possible for these approaches to fool human observers? Besides the quality of generated face images, other factors that might affect a human observer's decision when distinguishing between synthetic and real faces remain unclear.

 \begin{figure}[!t]
  \centering
  \begin{minipage}[b]{0.105\textwidth}
    \includegraphics[width=\textwidth]{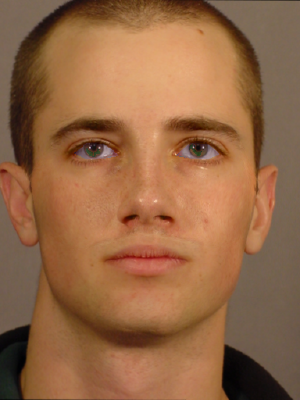}
  \end{minipage}
  \hfill
  \begin{minipage}[b]{0.105\textwidth}
    \includegraphics[width=\textwidth]{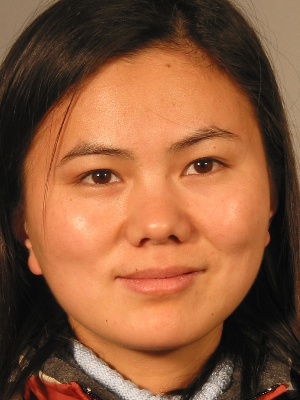}
  \end{minipage}
  \hfill
  \begin{minipage}[b]{0.14\textwidth}
    \includegraphics[width=\textwidth]{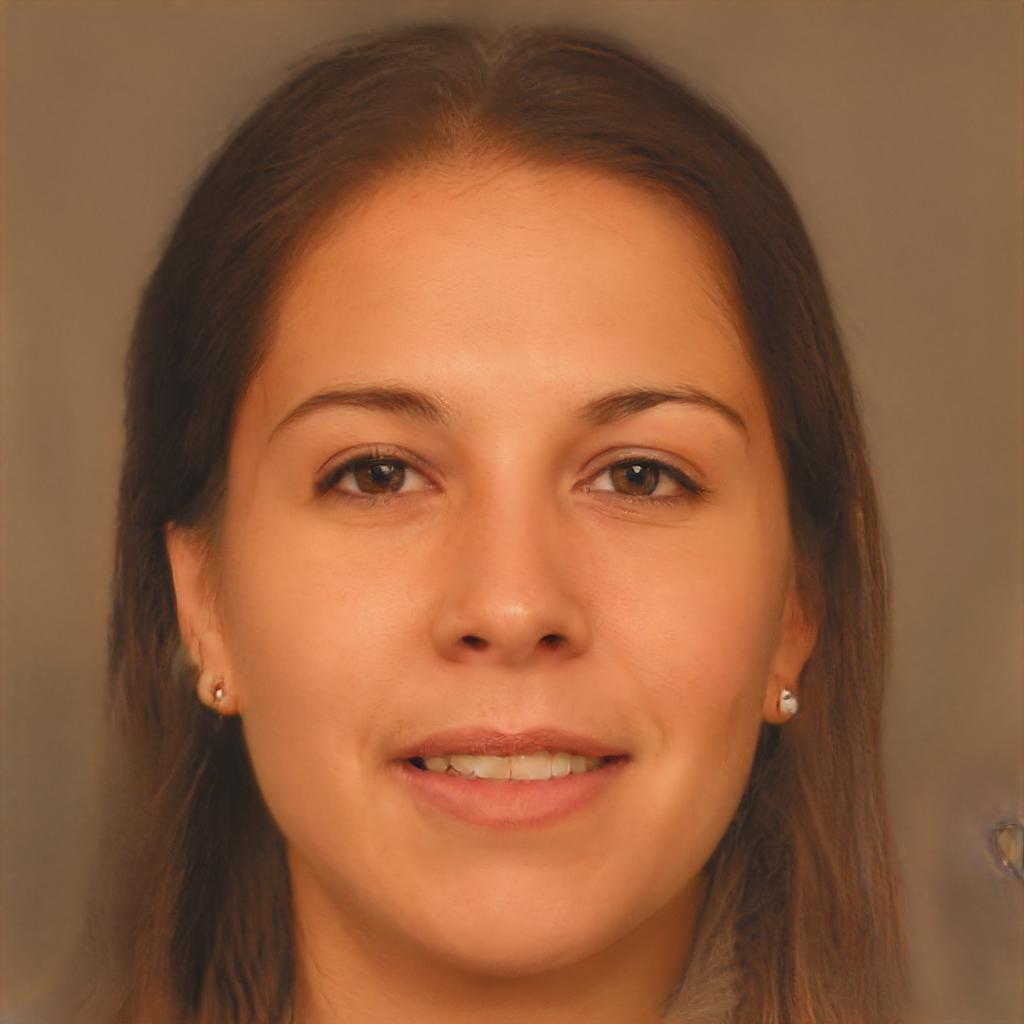}
  \end{minipage}
  \hfill
  \begin{minipage}[b]{0.105\textwidth}
    \includegraphics[width=\textwidth]{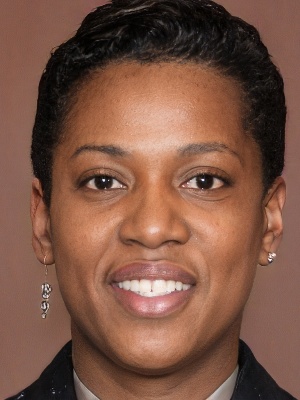}
  \end{minipage}
  \caption{Can you tell which face in the above set is from a real identity? The pace of face synthesis research has accelerated rapidly, and it is now possible for anyone to create photo-realistic faces of non-existent identities using accessible software tools. This has been good for creatives on the Internet, but potentially problematic for ordinary people who may encounter a fake face as part of a scam or some other deceptive practice. In this paper, we study the human ability to distinguish between real and synthetic faces.}
  \label{fig:teaser_fig}
  \vspace{-5mm}
\end{figure}


Motivated by these issues, in this work we studied the human perception of synthetic faces generated by different algorithms using methods from experimental psychology. The primary question we're interested in is: how well can humans distinguish between synthetic faces and real faces? This is of great importance because of the increasing use of synthetic faces in espionage~\cite{Satter_2019}, propaganda~\cite{Rawnsley_2020}, and online trolling~\cite{O'Sullivan_2020}. For example, such technology might bring severe consequences if deployed in a deceptive manner in video conferencing, as this communication tool is playing a key role in our lives during the COVID-19 pandemic. Websites such as \url{thispersondoesnotexist.com} have gained increasing attention, showing the power of GANs in creating synthetic identities with fine details. Datasets of synthetic face images and videos have also been released to accelerate the development of deep learning algorithms to detect this type of content~\cite{ciftci2020fakecatcher, korshunov2019vulnerability, rossler2018faceforensics, rossler2019faceforensics++}. 

There are several aspects of the visual appearance of the face that we are interested in assessing in this work. We know that synthetic faces look realistic at first glance, but can people distinguish them from real human faces when both are presented together? 
In Fig.~\ref{fig:teaser_fig}, only the face that is second from the left is from a real identity. The other face images are all generated by face synthesis techniques. The two images on the right are generated by StyleGAN~\cite{karras2019style}, while the leftmost image is generated by SREFI~\cite{banerjee2017srefi}. Identifying the real face from these images is an extremely difficult task. There are multiple reasons for that. 

First, the quality of synthetic faces has achieved a level whereby generated faces possess a significant number of high-quality details such as skin texture, hair, and wrinkles.
Irregularities in such details once served as clues that people would try to find when attempting to distinguish  artificial faces from real ones. Second, in the study introduced in this paper, human raters are presented with images of varying gender, race, and age, as in Fig.~\ref{fig:teaser_fig}, which makes the task even harder. However, this scenario mimics the deployment of synthetic faces in the real-world, where they can be mixed with real faces on social media and elsewhere. Third, the context in which a face appears is another confound. To what extent do overall scene conditions such as the background or lighting affect human judgement of real and synthetic faces? 

To understand the human perception of synthetic faces, we designed four behavioral experiments with different elements that could be controlled for (\textit{e.g.}, background, lighting, subject identities). Alarmingly, statistical analysis of the collected data shows that the human capacity to distinguish between real and synthetic faces is no better than chance at best. We also look at potential information leakage in synthetic faces.  Since synthetic faces are always generated using face images of real identities as part of the process, is it possible that a resulting synthetic face can reveal the real identity that contributed to it in some capacity? And will human observers notice?
 

This work offers a thorough and in-depth evaluation of the human perception of synthetic faces that is different from existing studies in several regards:

\begin{enumerate}
    \item Different face synthesis techniques are utilized including a state-of-the-art GAN-based method, which makes our analysis more comprehensive in scope. 
    \item Instead of relying on a single source for synthetic and real face images, a new large-scale dataset is created from multiple sources of face images, giving our study greater image diversity.
    \item The study of human perception of synthetic faces is grounded in established methods and procedures from experimental psychology, giving the analysis a level of rigor missing from prior studies.
    \item Large-scale crowdsourced behavioral experiments were conducted using hundreds of human subjects, making this one of the largest studies of its kind. 
\end{enumerate}

In the rest of this paper, we review the literature (Sec.~\ref{sec:related}) on traditional image processing-based face synthesis, current machine learning-based face synthesis, and existing studies of the human perception of synthetic faces. We then go on to discuss the sources of data used in our study (Sec.~\ref{sec:data}), as well as the experimental protocols and results (Sec.~\ref{sec:experiments}). We conclude with some thoughts on the landscape of synthetic media, and ideas for mitigating the impact of synthetic faces when they are used in questionable circumstances. 

%% file: 3background.tex
\section{Related Work}
\label{sec:related}
Face synthesis has been an active computer vision research area for decades. The main goal of face synthesis is to generate realistic new identities from images of real faces.

\subsection{Traditional face generation methods}
There are multiple ways to combine two or more real faces into a new face~\cite{bitouk2008face, mosaddegh2014photorealistic, datta2012two}.  Bitouk~\cite{bitouk2008face} et al. proposed a system for face swapping and created a large library of face images that served as a pool of candidates for swapping operations. To swap a target face, the system picks candidates that most resemble the target face and adjusts them in pose, lighting, and color to match the input. After blending the adjusted faces into the target face, the system will rank the output images generated by each candidate via a boundary metric and pick the best one. The boundary metric measures the perceptual similarity to the input image along the boundary of the replacement region.
A user study suggested that the generated synthetic faces are frequently mistakenly identified as real.

Mosaddegh et al.~\cite{mosaddegh2014photorealistic} defined different regions, such as the eyes, eyebrows, and nose of faces as face components. To synthesize a new face from an input face, they borrow different face components from up to four donor faces and replace the respective regions in the original input face image with these borrowed components.

Banerjee et al.~\cite{banerjee2017srefi} proposed a similar approach dubbed ``synthesis  of  realistic  example  face images'' (SREFI), which constructs synthetic face images from a set of face images from real identities. SREFI triangularizes faces, while keeping important facial features away from triangle corners, and then stitches and blends together triangles combined from a set of different face images.

Other proposed methods have involved embedding faces into high-dimensional spaces, allowing faces to be synthesized by traversing the resulting manifold~\cite{huang2009manifold}.

\subsection{Machine learning-based face generation}
Compared to the traditional methods, machine learning-based techniques that make use of deep neural networks show significant advantages in image quality and flexibility of use. 
Among the various deep neural networks in the literature, GANs have shown their superiority in generating realistic objects from different inputs~\cite{isola2017image}. 

Leveraging this early success, GANs have been successfully applied to the specific object category of human faces. Song et al.~\cite{song2018geometry} proposed a geometry-guided GAN for photo-realistic and identity-preserving facial expression synthesis. Wang et al.~\cite{wang2018facial} proposed a novel U-Net Conditional GAN and an identity-preserving loss for facial expression generation. Antipov et al.~\cite{antipov2017face} proposed a GAN-based method for automatic identity-preserving face aging. Zhang et al.~\cite{zhang2019synthesis} were able to use a GAN-based multi-stream feature-level fusion technique to generate high-quality face images from the input of polarimetric thermal images. To solve the problem of identifying faces with significant out-of-plane rotation, Han et al.~\cite{han2019face} proposed a GAN with tripartite adversaries to synthesize the frontal view from various poses. 

Beyond generating new instances of the same identity, GANs have been widely used to generate `random' faces for new synthetic identities. Yuan et al.~\cite{yuan2021attributes} proposed an attribute aware face generator with a GAN that can generate face images according to specific characteristics corresponding to given attributes.
StyleGAN~\cite{karras2019style} can disentangle the latent space of the styles of face images (\textit{e.g.}, hair styles, face shapes, eye colors), therefore providing a more controlled process of face synthesis.


\subsection{Studies on the human perception of synthetic faces}
Studies of the human perception of synthetic faces have been conducted ever since computer graphics matured to the point of being able to generate photo-realistic objects. For example, Farid and Bravo~\cite{farid2012perceptual} conducted psychological experiments that explored the impact of face image resolution, compression, and color on the human ability to distinguish synthetic faces from real face images.
Balas and Pacella~\cite{balas2017trustworthiness} studied the difference in the perception of trustworthiness in real faces and synthetic faces. Their results showed that the absolute levels of perceived trustworthiness and relative trustworthiness judgements are both affected by synthetic faces. The same group also conducted experiments related to memory during human interactions with artificial faces~\cite{balas2015artificial}. Results indicated it is easier to remember real faces. 

After the emergence of high-quality synthetic face generation powered by deep neural networks, more studies appeared that focused on both human perception and the algorithms related to the face generation methods. 
Korshunov and Marcel~\cite{korshunov2020deepfake} conducted a study of the human perception of deepfake videos. Human raters were asked a question about whether the face in a video is real or synthetic. Specifically,  human perception was compared with the state-of-the-art deepfake detection algorithms. In this context, deepfakes are edited videos where one person's face is inserted onto the body of another person. The deepfake videos are split into different scales in terms of difficulty. The results show that there is a difference between human detection ability and deepfake detection algorithms, and that deepfake videos are able to confuse the majority of the public at the present moment. Lago et al.~\cite{Lago_2021} conducted a similar study on synthetic face images via crowdsourced experiments.

%% file: 4method.tex
\section{Data sources for Real and Synthetic Faces}
\label{sec:data}
In this work we analyzed human behavior in distinguishing between synthetic faces and real faces. To the best of our knowledge, this is the first study that quantifies the quality of synthetically-generated faces in real-world settings with respect to plausibility.

There are three types of face images used in our experiments: real faces, synthetic faces generated by SREFI~\cite{banerjee2017srefi}, and synthetic faces generated by StyleGAN2~\cite{karras2020analyzing}.

\subsubsection{Real faces}
\begin{figure}[!t]
  \vspace{-5mm}
  \centering
  \begin{minipage}[b]{0.155\textwidth}
    \includegraphics[width=\textwidth]{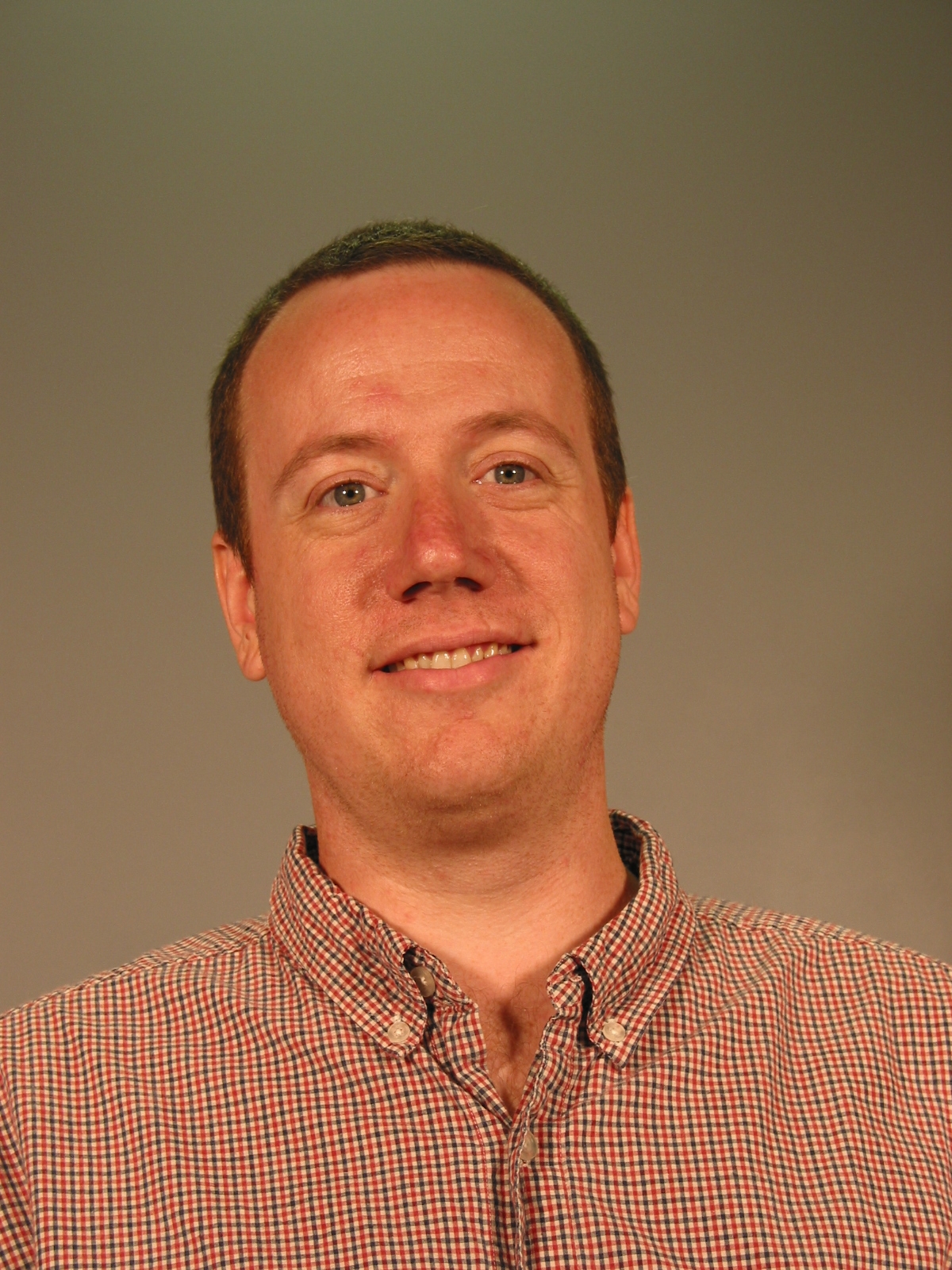}
  \end{minipage}
  \hfill
  \begin{minipage}[b]{0.155\textwidth}
    \includegraphics[width=\textwidth]{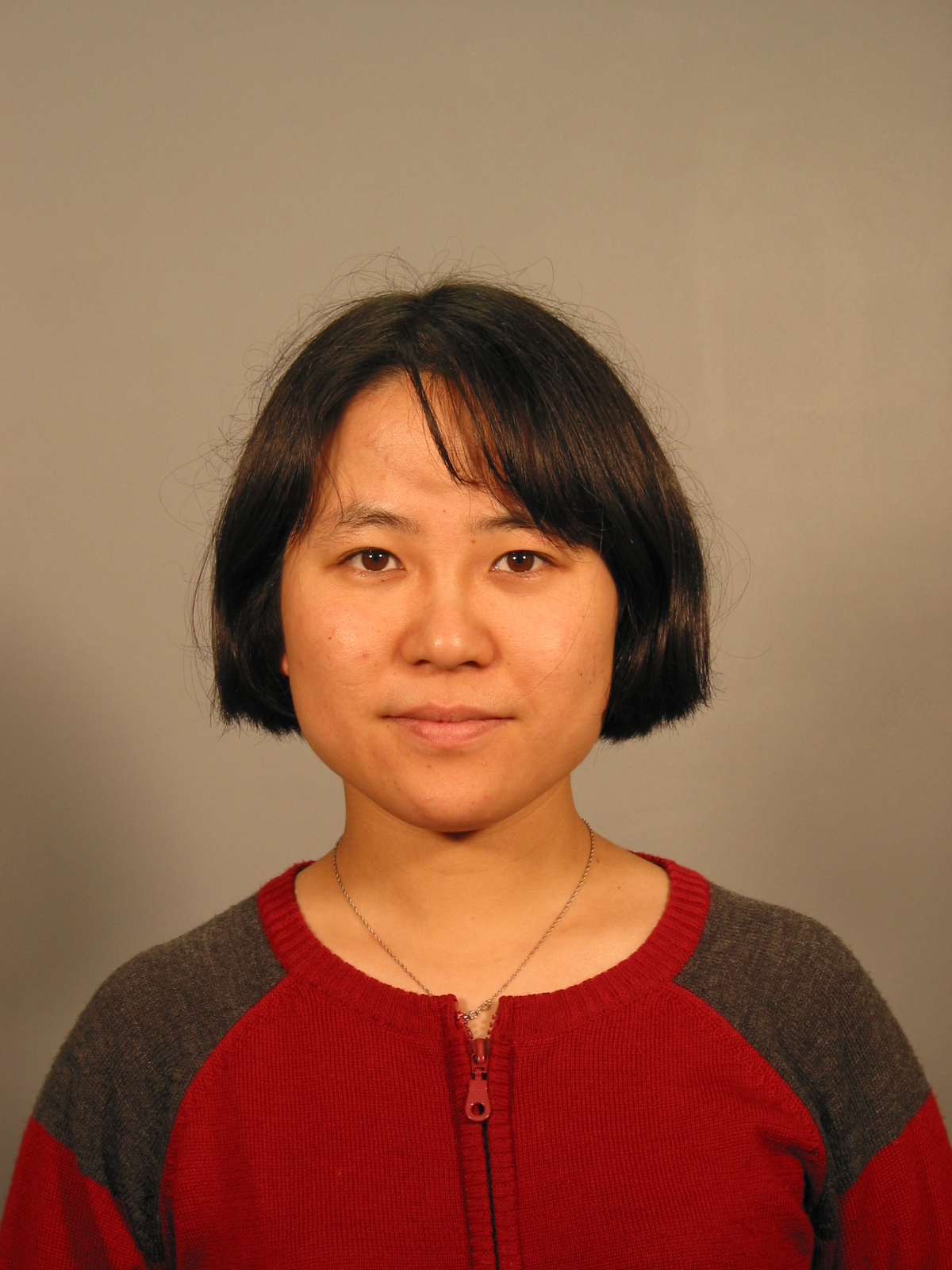}
  \end{minipage}
  \hfill
  \begin{minipage}[b]{0.155\textwidth}
    \includegraphics[width=\textwidth]{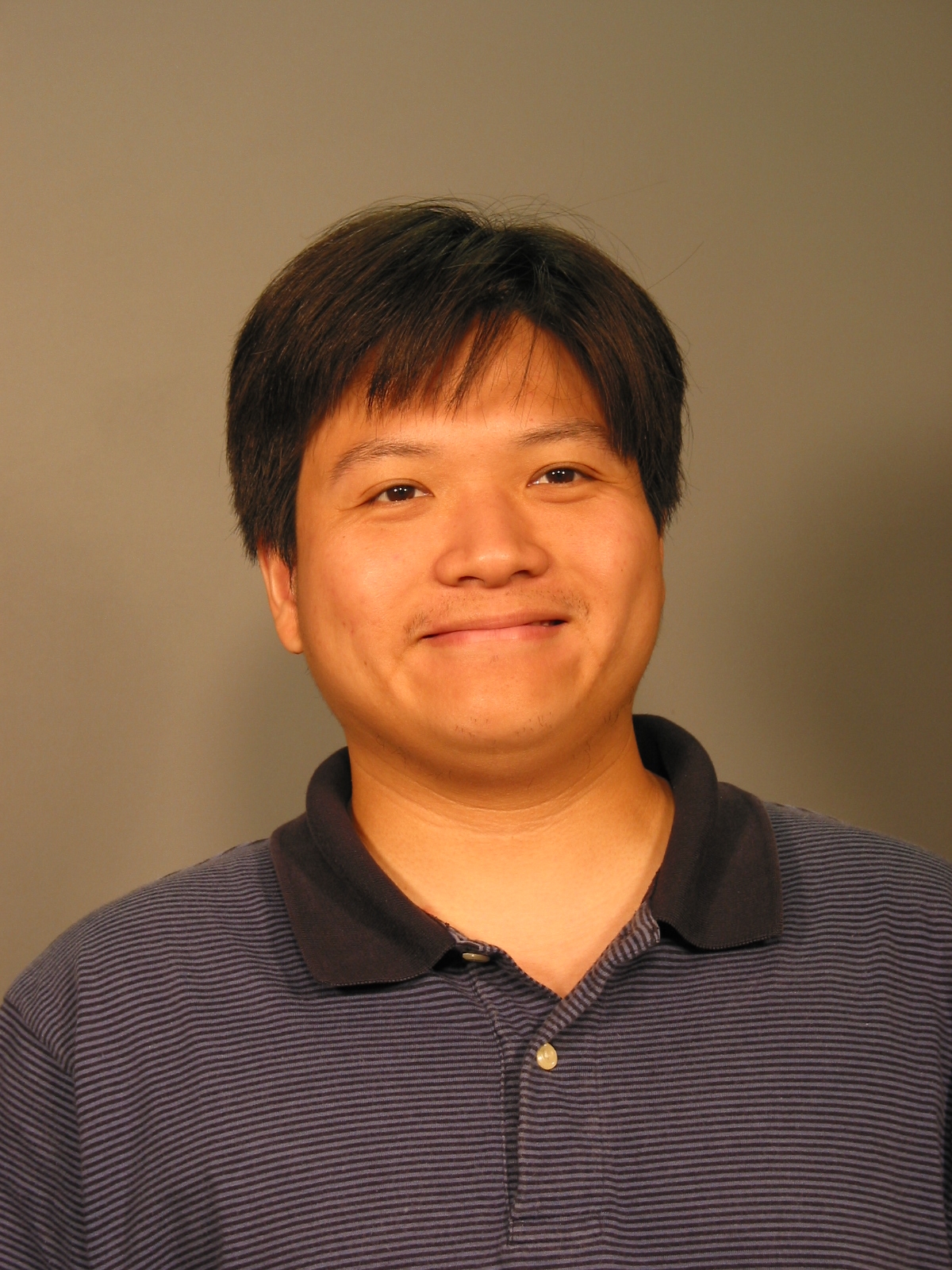}
  \end{minipage}
  \caption{Examples of real faces from the public dataset collected by Phillips et al.~\cite{phillips2017lessons}. Frontal face portraits of subjects with a neutral or happy emotion are used as the real face images in our experiments.}
  \label{fig:real_sample}
  \vspace{-5mm}
\end{figure}

\begin{figure}
  \vspace{-5mm}
    \centering
    \begin{minipage}[b]{0.235\textwidth}
    \includegraphics[width=\textwidth]{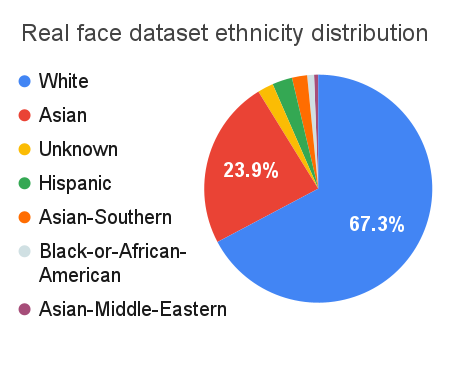}
    \end{minipage}
        \begin{minipage}[b]{0.235\textwidth}
    \includegraphics[width=\textwidth]{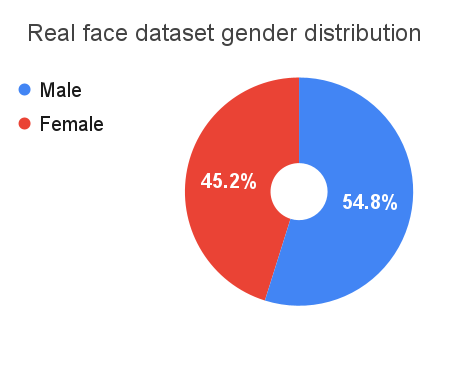}
    \end{minipage}
    \caption{Demographics of the subjects in the subset of the dataset of real faces~\cite{phillips2017lessons}  used in our experiments. There is a diversity of ethnicity and gender is close to balanced.}
    \label{fig:race_distribution}
    \vspace{-5mm}
\end{figure}

We used a publicly-available dataset collected by Phillips et al.~\cite{phillips2017lessons} as our source of real faces from which multiple frontal face images with different expressions were collected. The subjects varied in gender, age and ethnicity. Some sample faces are shown in Fig~\ref{fig:real_sample}. We used a subset of 1,023 images as a pool of real faces (RPool1) for our experiments. The ethnicity and gender distributions for this subset are shown in Fig.~\ref{fig:race_distribution}. Although the majority of the subjects are White and Asian,  a diversity of ethnicity is reflected in the subjects of our subset. The gender distribution is also close to being balanced. It is not exactly balanced because of the image quality requirements of the algorithms (some images have to be filtered out because they are unsuitable to use). The age of the subjects ranges from 18 years old to 69 years old.

\subsubsection{Synthetic faces generated by SREFI}
SREFI~\cite{banerjee2017srefi} is a face synthesis approach based on a convolutional neural network (CNN) face representation for associating similar face candidates, and it can construct an arbitrarily large number of synthetic faces from real identities. To synthesize a face, the input real face image is firstly triangularized into region-specific triangles, while important face areas are prevented from being placed in the corner of triangles. A set of ``donor'' faces (also real identities) that are close to each other in the CNN feature representation space are then selected. A VGG-Face CNN model~\cite{parkhi2015deep} pre-trained on the VGG-Face dataset~\cite{parkhi2015deep} is used to calculate features. All donor faces are triangularized like the base face.

\begin{figure}[b]
  \vspace{-5mm}
  \centering
  \begin{minipage}[b]{0.155\textwidth}
    \includegraphics[width=\textwidth]{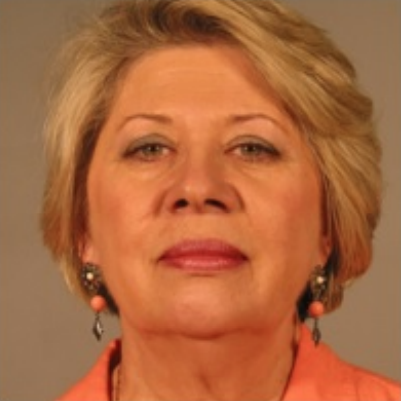}
  \end{minipage}
  \hfill
  \begin{minipage}[b]{0.155\textwidth}
    \includegraphics[width=\textwidth]{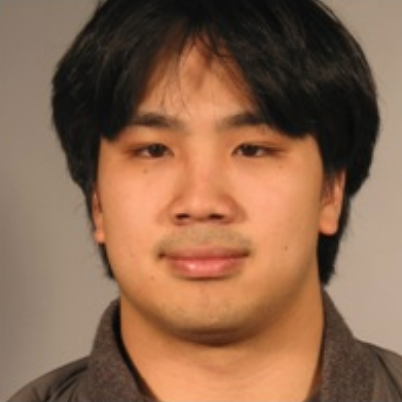}
  \end{minipage}
  \hfill
  \begin{minipage}[b]{0.155\textwidth}
    \includegraphics[width=\textwidth]{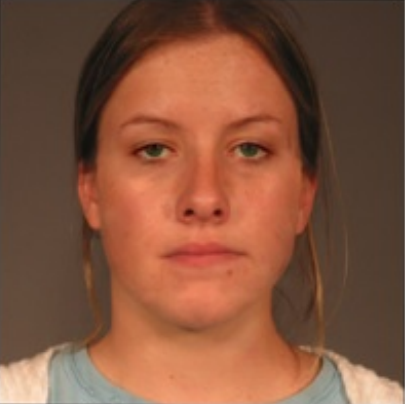}
  \end{minipage}
  \caption{Example of synthetic face generated using SREFI~\cite{banerjee2017srefi}. From left to right: donor face, donor face, a synthetic face generated using the two donor faces.}
  \label{fig:srefi_sample}
\end{figure}

Synthetic faces are generated by stitching together the corresponding triangles from the ``donor'' faces. To ensure the uniformity of the created face, vital facial regions including the mouth, nose and eyes are restricted so that they come from the same donor. A synthetic face is generated from a donor set with a size of seven to ten real face images. In this work, we created a synthetic face pool (SPool1) of 1,000 synthetic faces using SREFI. The donor faces come from the same public dataset generating RPool1. Example images generated using this method are shown in Fig.~\ref{fig:srefi_sample}.


\subsubsection{Synthetic faces generated by GANs}
As GANs are very successful in generating high-quality synthetic face images, there are multiple websites releasing the outputs of these networks to the public for any use they see fit. We collected synthetic faces generated by GANs by downloading images from the website \url{thispersondoesnotexist.com}. 

\begin{figure}[!t]
  \vspace{-5mm}
  \centering
  \begin{minipage}[b]{0.155\textwidth}
    \includegraphics[width=\textwidth]{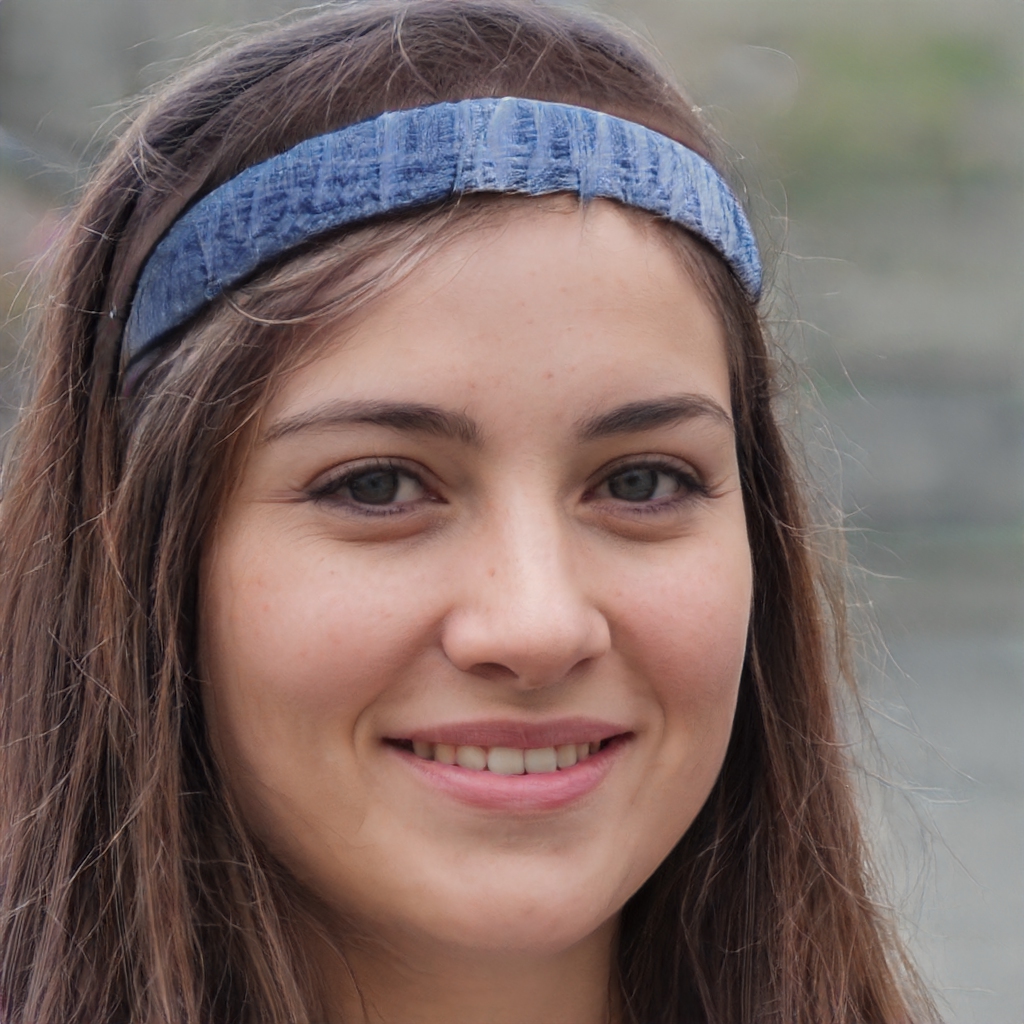}
  \end{minipage}
  \hfill
  \begin{minipage}[b]{0.155\textwidth}
    \includegraphics[width=\textwidth]{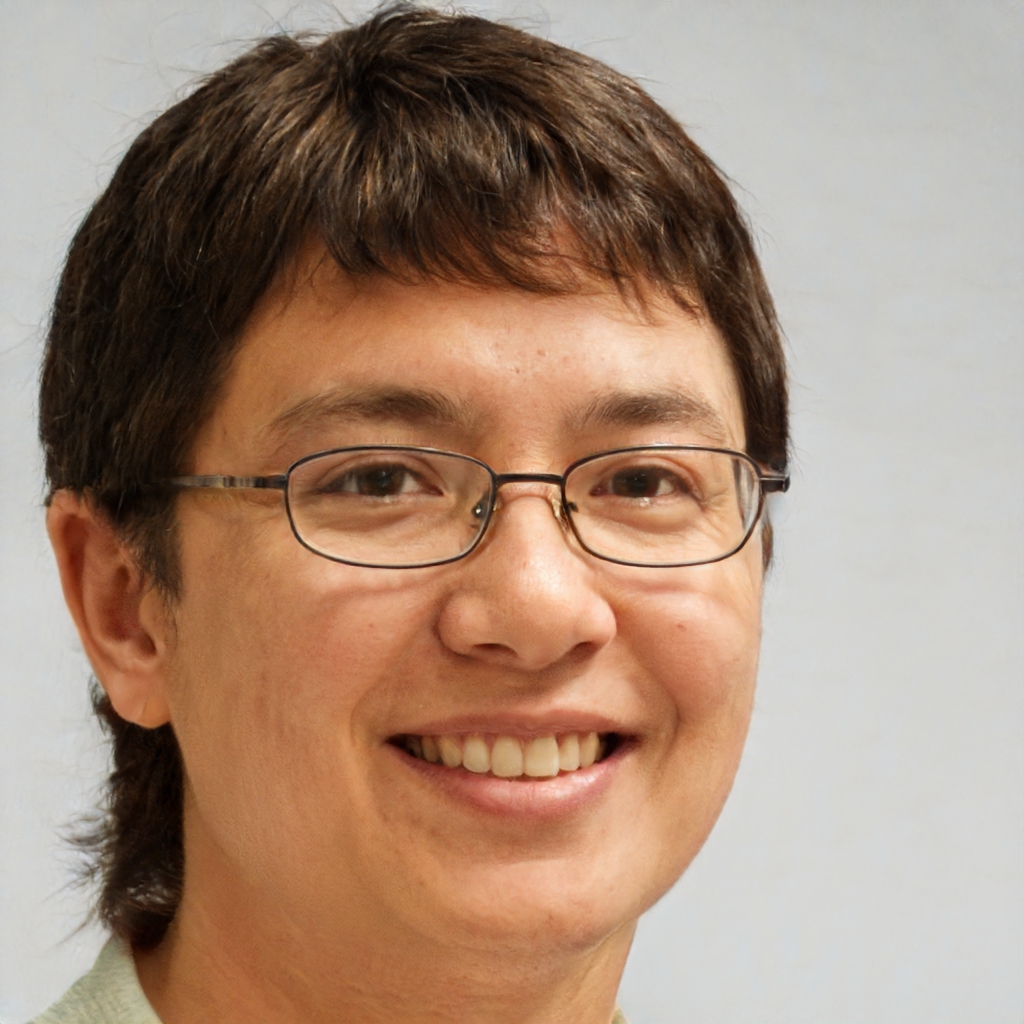}
  \end{minipage}
  \hfill
  \begin{minipage}[b]{0.155\textwidth}
    \includegraphics[width=\textwidth]{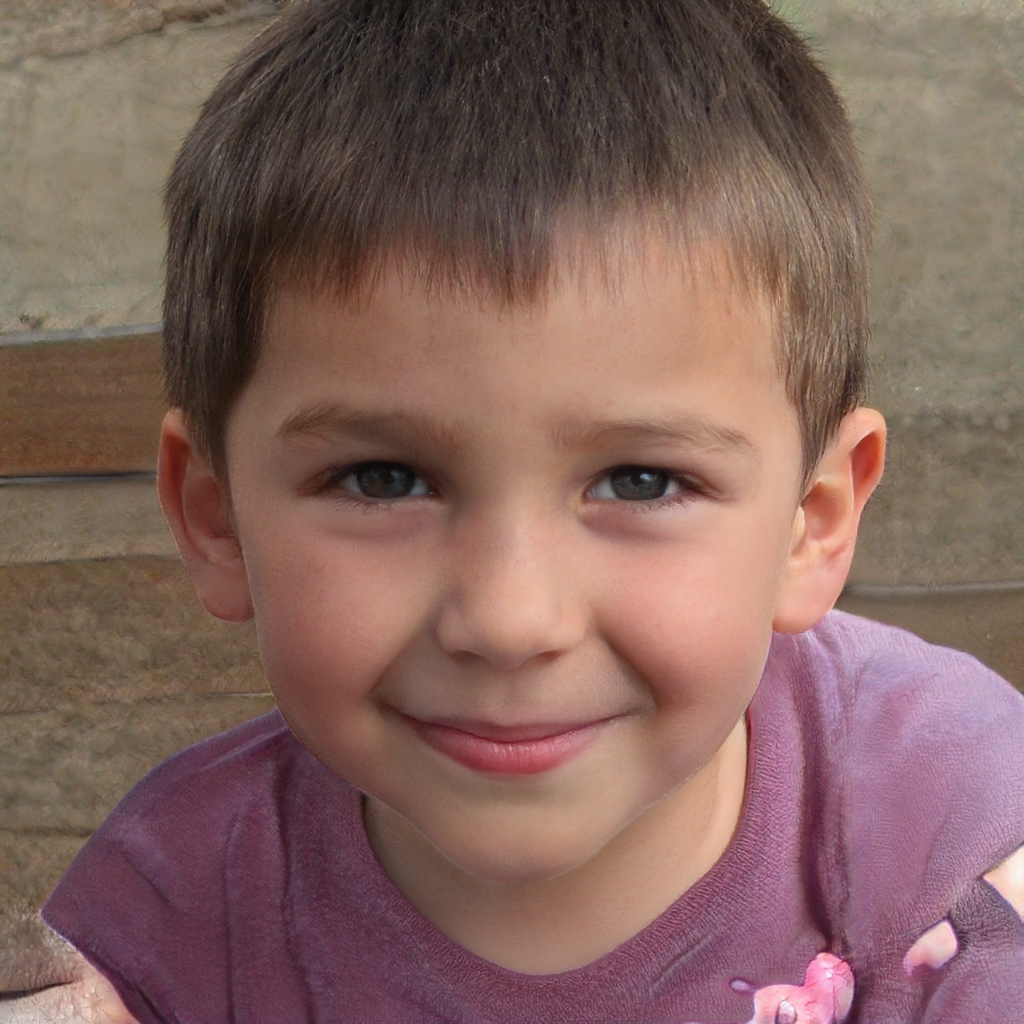}
  \end{minipage}
  \caption{Examples of synthetic faces generated using StyleGAN2~\cite{karras2020analyzing}. GAN images used for our experiments  are collected from \url{thispersondoesnotexist.com}.}
  \label{fig:gans_sample}
  \vspace{-5mm}
\end{figure}

All of the images in this set are generated using StyleGAN2~\cite{karras2020analyzing}. StyleGAN~\cite{karras2019style} is a style-based generator that adjusts the style of each convolutional layer's output based on the input latent code, therefore directly controlling the face synthesis process. Lower resolutions control high-level styles of the output faces such as pose and face shape. A block of layers in the middle of the generator controls the generation of facial features such as the eyes and expression. Layers closer to the output layers control the fine details, such as the color scheme. Using StyleGAN it is possible to generate high-resolution (1024$\times$1024) synthetic face images with stochastic variation (\textit{e.g.}, freckles, hair). StyleGAN2 proposed certain modifications to further improve the image quality of StyleGAN's output. The improvements to the generator's architecture include weight demodulation, lazy regularization and path length regularization. StyleGAN2 also revisits the progressive growth idea to stabilize the training process. Overall, StyleGAN2 is able to generate images with even finer details after all these changes.
Some of the synthetic GAN-generated faces that were collected are shown in Fig~\ref{fig:gans_sample}. The total number of images in this synthetic face pool (SPool2) is 1,500.

A summary of the data sources is shown in Table \ref{table: datasource}.
\begin{table}[h!]
  \vspace{-2mm}
\centering
\caption{Sources of face images used in our experiments.}

\begin{tabular}{ |c |c |c |c |}
\hline
 & RPooL1 & SPooL1 & SPooL2 \\ 
\hline
 Number of images & 1,023 & 1,000 & 1,500   \\
 \hline
\end{tabular}
  \vspace{-5mm}

\label{table: datasource}
\end{table}

%% file: 5experiment.tex
\section{Human Perception Experiments}
\label{sec:experiments}

We conducted four independent experiments with human raters to examine the plausibility of synthetic faces generated using different methods. 
%
%
The first three experiments aim at investigating whether the synthetic faces will lead to failure in people distinguishing between real and synthetic faces and how environmental context affects judgement. The last experiment was designed to shed light on the question of how much of the identity signal from real faces used as input to a face generation approach leaks into synthetic faces.

A two-alternative forced-choice (2AFC) procedure is used in all four experiments. This means the observer is presented with a pair of visual stimuli (\textit{e.g.}, a pair of face images), and then has to select one of two choices (\textit{e.g.,} ``left" or ``right"). Since answers outside of the provided two options are not allowed, the observer's choice is forced between the two alternatives. The 2AFC design allows scientists to measure the statistics for each visual stimulus pair across all observers and interpret the results.

\begin{figure}[t]
  \vspace{-5mm}
  \centering
  \begin{minipage}[b]{0.116\textwidth}
    \includegraphics[width=\textwidth]{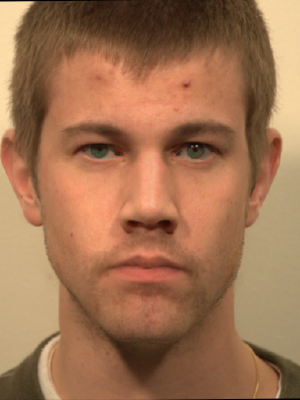}
  \end{minipage}
  \hfill
  \begin{minipage}[b]{0.116\textwidth}
    \includegraphics[width=\textwidth]{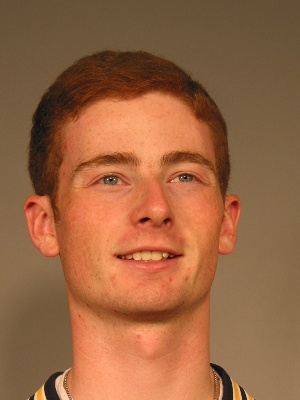}
  \end{minipage}
  \hfill
  \begin{minipage}[b]{0.116\textwidth}
    \includegraphics[width=\textwidth]{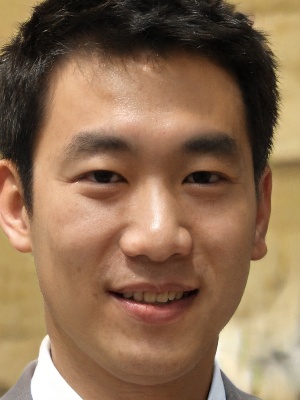}
  \end{minipage}
    \hfill
  \begin{minipage}[b]{0.116\textwidth}
    \includegraphics[width=\textwidth]{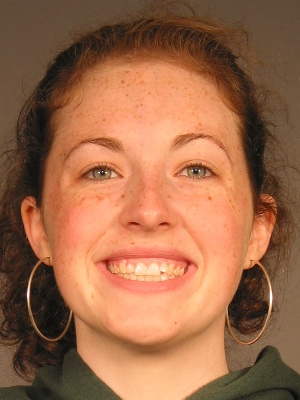}
  \end{minipage}
  \caption{Two sample pairs from Experiment 1. From left to right: synthetic face in Pair 1, generated by SREFI; real face in Pair 1; synthetic face in Pair 2, generated by StyleGAN2; real face in Pair 2.}
  \label{fig:p1_sample}
  \vspace{-5mm}
\end{figure}

\begin{figure}[b!]
  \vspace{-5mm}
  \centering
  \includegraphics[width=0.48\textwidth]{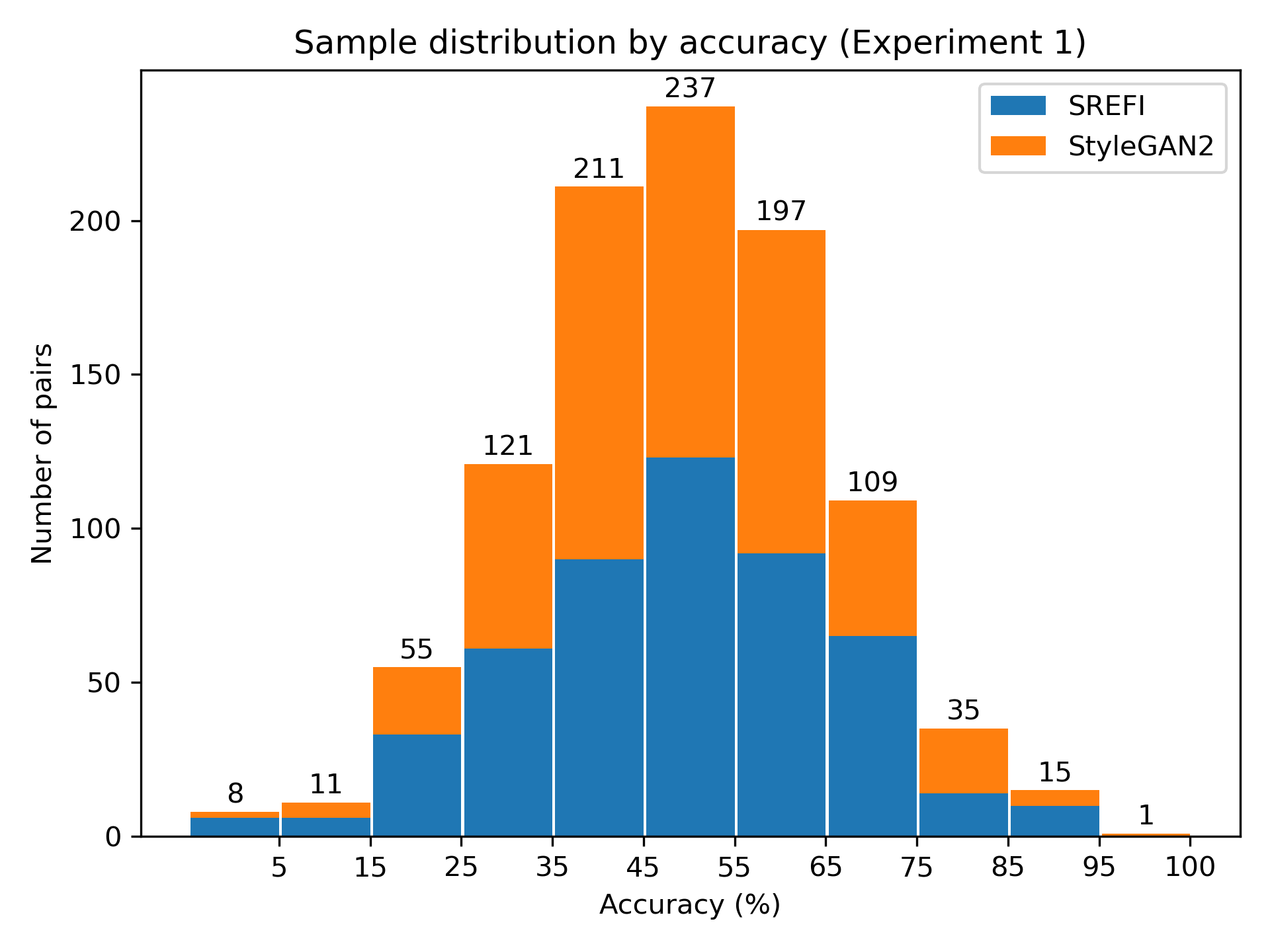}
  \caption{Human performance on 1,000 pairs of real / synthetic face images with the background visible. Each bar represents the number of pairs that has an accuracy larger than the lower bound and smaller than or equal to the upper bound. For example, the bar between 45\% and 55\% reflects that there are 237 pairs of images that have an accuracy greater than 45\% while being less than or equal to 55\%. }
  \label{fig:exp01_result}
\end{figure}

\subsection{Experiments 1-3: Which face is real or synthetic?}

\subsubsection{Distinguishing faces with full context}
In this experiment, we randomly selected 1,000 real face images from RPool1 and 500 synthetic face images from both SPool1 and from SPool2, which resulted in 1,000 synthetic face images in total. Each real face was randomly paired with a synthetic face, which resulted in 1,000 real / synthetic face image pairs. All face images are pre-processed in the same manner. We first detect face bounding boxes using Haar feature-based cascade classifiers. Then the images are cropped with an aspect ratio of 3:4 at the center of the detected face bounding box. Finally, we resize the cropped face images to 300$\times$400 in resolution, such as the samples shown in Fig.~\ref{fig:p1_sample}. Human subjects are presented with one pair of face images side by side in each trial and asked a question of ``which of the faces is real (or synthetic)?", where the target question is varied randomly. Subjects are prompted to choose ``left" or ``right" as an answer and the images stay on the screen until the choice has been made.

\begin{figure}[!t]
  \vspace{-5mm}
  \centering
  \begin{minipage}[b]{0.116\textwidth}
    \includegraphics[width=\textwidth]{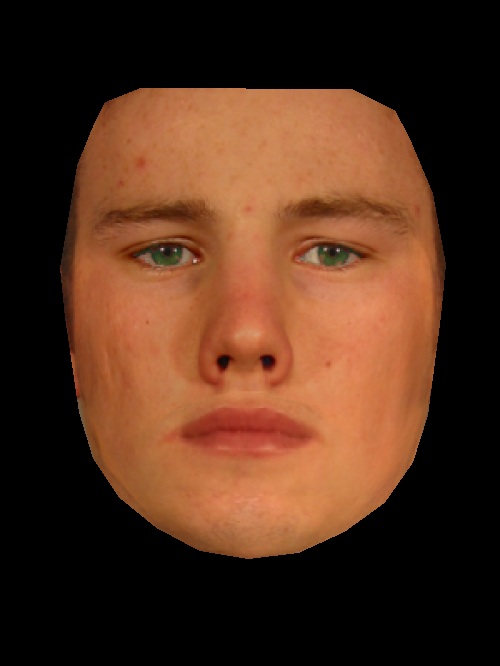}
  \end{minipage}
  \hfill
  \begin{minipage}[b]{0.116\textwidth}
    \includegraphics[width=\textwidth]{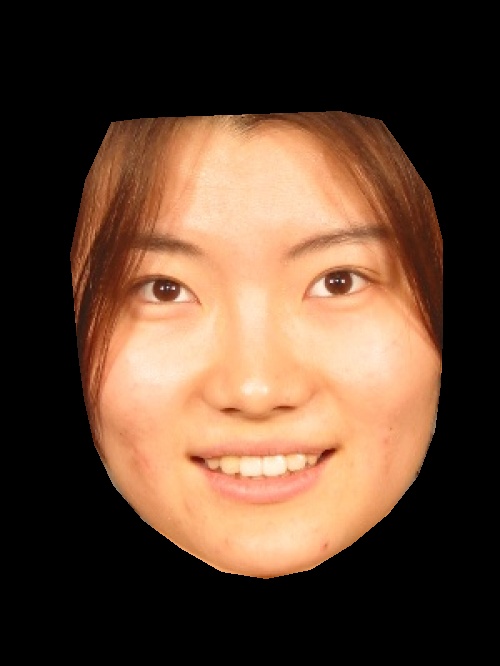}
  \end{minipage}
  \hfill
  \begin{minipage}[b]{0.116\textwidth}
    \includegraphics[width=\textwidth]{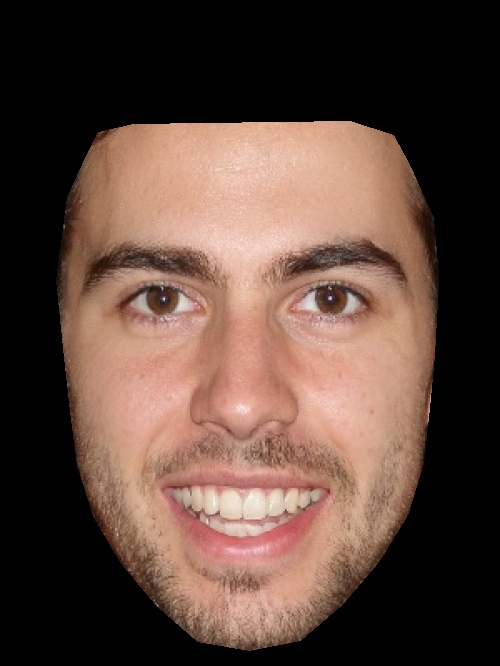}
  \end{minipage}
    \hfill
  \begin{minipage}[b]{0.116\textwidth}
    \includegraphics[width=\textwidth]{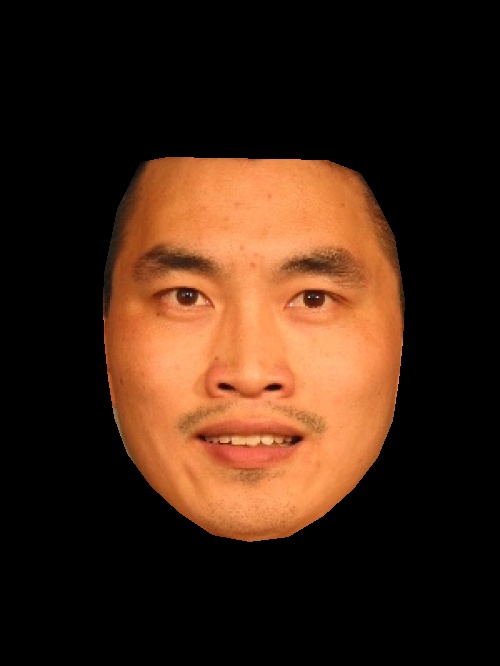}
  \end{minipage}
  \caption{Two sample pairs from Experiment 2. From left to right: synthetic face in Pair 1, generated by SREFI; real face in Pair 1; synthetic face in Pair 2, generated by StyleGAN; real face in Pair 2.}
  \label{fig:p2_sample}
  \vspace{-5mm}
\end{figure}

There are 176 human raters involved in this experiment, which was crowdsourced on Amazon's Mechanical Turk Service. For experiments 1-3, workers were paid \$5 to rate 50 pairs of images. For each real / synthetic face image pair, at least 10 workers are assigned to answer the question. On average 10.75 people viewed each pair of images. 
We calculated the accuracy for each pair and the distribution across pairs is shown in Fig.~\ref{fig:exp01_result}. Here accuracy is the number of correct responses (\textit{e.g.}, ``left" was selected when the real face image was placed on the left and the subject was asked ``which image is real?") divided by the total number of responses collected for this specific pair. 

Considering Fig.~\ref{fig:exp01_result}, the data appear to be normally distributed. This brings into question the human ability to accurately discriminate between real and synthetic faces. To investigate this, we performed a one-tailed one sample t-test on these data. The null hypothesis states that human raters are no better than chance at identifying synthetic faces: $H_{0}: Acc_{human} = 0.5$. Two alternative hypotheses can also be formulated in the one-tailed scenario. The alternative hypothesis considering the left tail states that human raters tend to mistake synthetic faces as real faces (worse than chance): $H_{a_1}: Acc_{human} < 0.5$. The alternative hypothesis considering the right tail states that human raters tend to identify synthetic faces correctly (greater than chance): $H_{a_2}: Acc_{human} > 0.5$. 

As shown in Table~\ref{table: exp_result}, for the test assessing the right tail, the t-statistic is -1.834 and $p$ is greater than 0.05 (the 95\% confidence interval). Thus the null hypothesis is not rejected. Given the distribution in Fig.~\ref{fig:exp01_result}, this isn't a surprising finding. What is more interesting is that for the left tail, $p$ is less than 0.05, meaning the null hypothesis is rejected: people tend to mistake synthetic faces as real faces in this experiment. Why might that be the case?

There are a few possibilities as to why this occurs. The first two are related to the facial features: the region of skin just below the eyes and the horizontal line-up of the eyes with the frame of the face.  Many of the synthetic images have quite smooth skin under the eyes, and the eyes are lined up on the exact horizontal. In essence, the synthetic faces, with smooth skin and good symmetry, may look more pleasing to human observers. There is also the possibility that the background is attracting undue attention, causing people to make incorrect decisions based on irrelevant information. The next experiment, which removes the backgrounds, will help us assess these possibilities. 


\begin{table}[b]
  \vspace{-5mm}
\centering
\caption{Statistics for Results of Experiments 1-4.}
\begin{tabular}{ |c |c |c |c |c |c |}
\hline
& \multirow{2}{*}{Mean} & \multirow{2}{*}{Std.} & \multicolumn{3}{c|}{One-tailed One Sample t-test}\\
\cline{4-6}
&  &  & t-stat & p-value (left)&  p-value (right)\\
\hline
Exp. 1  & 0.491 & 0.161 & -1.834 & \textbf{0.033}   & 0.967\\
\hline
Exp. 2  & 0.497 & 0.160 & -0.675 & 0.250  & 0.750\\
\hline
Exp. 3  & 0.497 & 0.163 & -0.626 & 0.266   & 0.734\\
\hline
Exp. 4  & 0.501 & 0.183 & 0.133 & 0.553 & 0.447\\
\hline
\end{tabular}
\label{table: exp_result}
\end{table}

\begin{table}[b]
\centering
\caption{Statistics for Results of Experiments 1-4.}
\begin{tabular}{ |c |c |c |}
\hline
&Mean& Std.\\
\hline
Real Same Identities  & 0.503 & 0.191\\
\hline
Real Different Identities  & 0.522 & 0.166\\
\hline
Real SREFI Faces & 0.491 & 0.185\\
\hline
Real StyleGAN Faces & 0.507 & 0.173\\
\hline
\hline
Overall & 0.501 & 0.133\\
\hline

\end{tabular}
\label{table: exp_result}
\end{table}

\subsubsection{Distinguishing faces with only the face region visible}
To examine whether the background information (\textit{e.g.} scene, hair, clothes, and pose) helps people recognize a synthetic face or  interferes with their judgement, we conducted a second experiment using Mechanical Turk.
1,000 real / synthetic face pairs were generated as in the previous experiment. However, in the pre-processing step, images are further segmented to only keep the face region. The segmentation is performed by using a pre-trained 81 landmark facial alignment model offered by Dlib~\cite{dlib09}, which made use of Histogram of Oriented Gradients (HOG) features combined with a linear classifier. Important face components including the forehead are preserved. Some sample images are shown in Fig.~\ref{fig:p2_sample}.  As in Experiment 1, raters are asked ``which face in the pair is real (or synthetic)?''
and images are left on the screen until the choice has been made. 

\begin{figure}[t]
  \vspace{-5mm}
  \centering
  \includegraphics[width=0.48\textwidth]{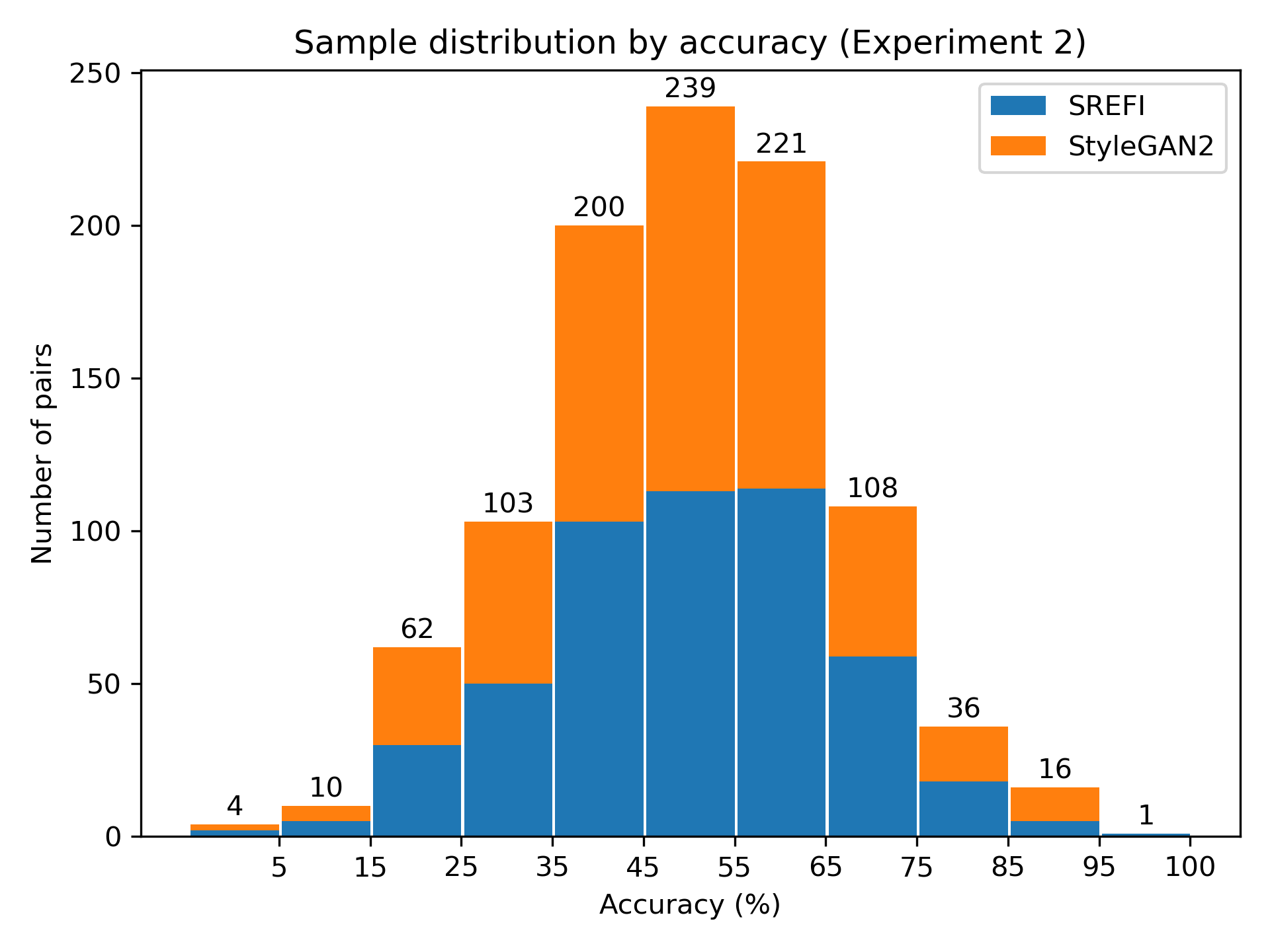}
  \caption{Human performance on 1,000 pairs of real / synthetic face images with only the face area visible.}
  \label{fig:exp02_result}
  \vspace{-5mm}
\end{figure}

For each real / synthetic pair of images, we assigned at least 10 people to answer the question. We collected responses for 10,550 pairs, and 174 different human raters were involved. Each pair of face images has an average of 10.55 rater responses. The accuracy distribution is shown in Fig.~\ref{fig:exp02_result}, which once again appears to be normally distributed. 

We performed a one-tailed one sample t-test on the result as in Experiment 1, where the hypotheses remained the same. As shown in Table~\ref{table: exp_result}, the p-values indicate that the null hypothesis cannot be rejected at the 95\% confidence interval. Therefore human performance is not better  than randomly guessing when recognizing synthetic faces by just the face region. Combined with the results from Experiment 1, this suggests that synthetic faces have a higher chance of fooling people with a background. Some people are likely relying on details (\textit{e.g.} hair, clothes, and scene) outside the face region in determining whether a face is synthetic or real, meaning smooth skin and symmetry are not significant clues. Current face synthesis techniques appear to be quite good at producing realistic contextual details in these images. 


 \begin{figure}[b]
  \vspace{-3mm}
  \centering
    \hfill
  \begin{minipage}[b]{0.116\textwidth}
    \includegraphics[width=\textwidth]{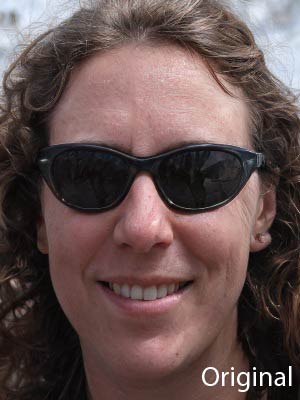}
  \end{minipage}
  \begin{minipage}[b]{0.116\textwidth}
    \includegraphics[width=\textwidth]{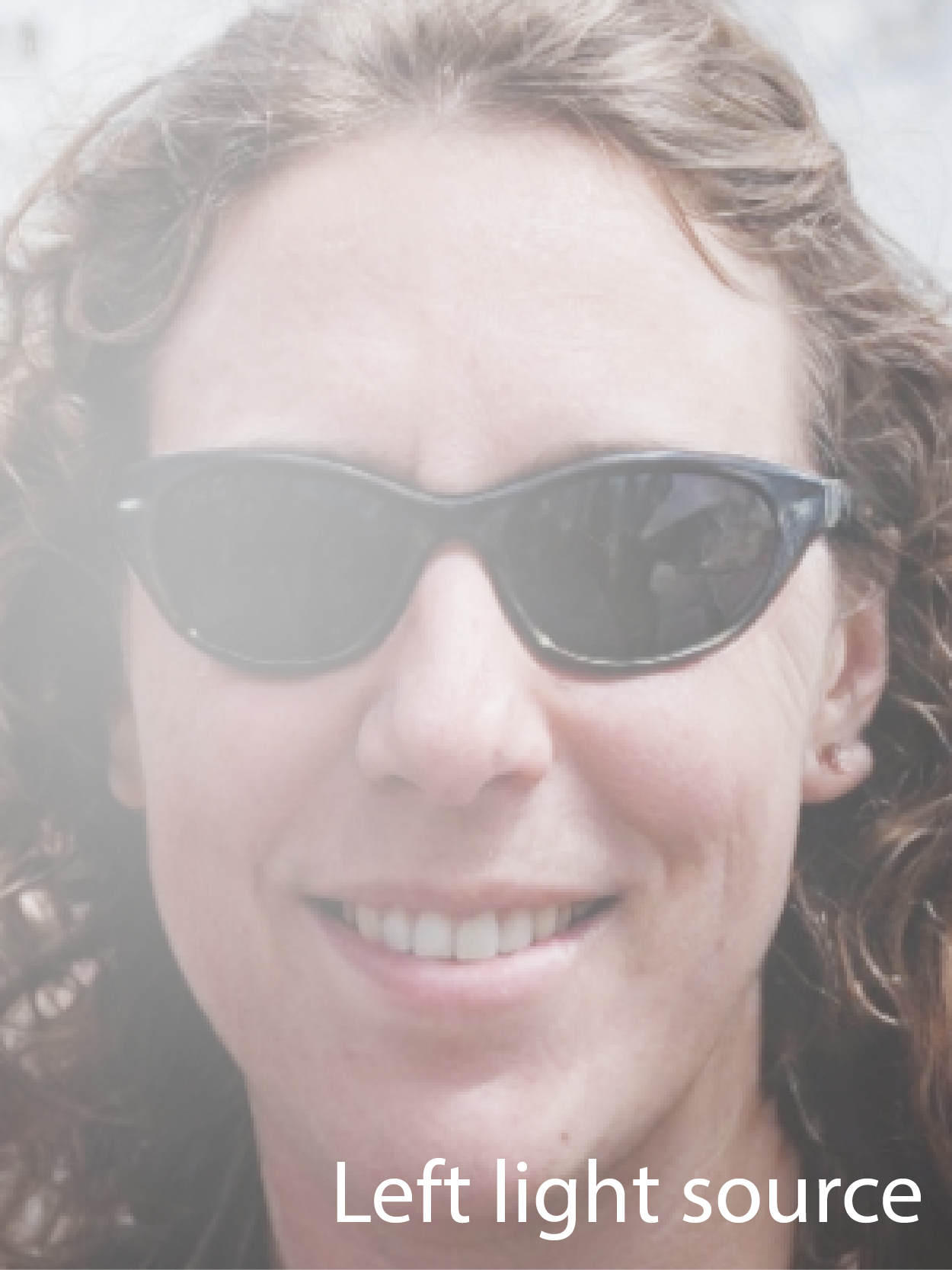}
  \end{minipage}
  \hfill
  \begin{minipage}[b]{0.116\textwidth}
    \includegraphics[width=\textwidth]{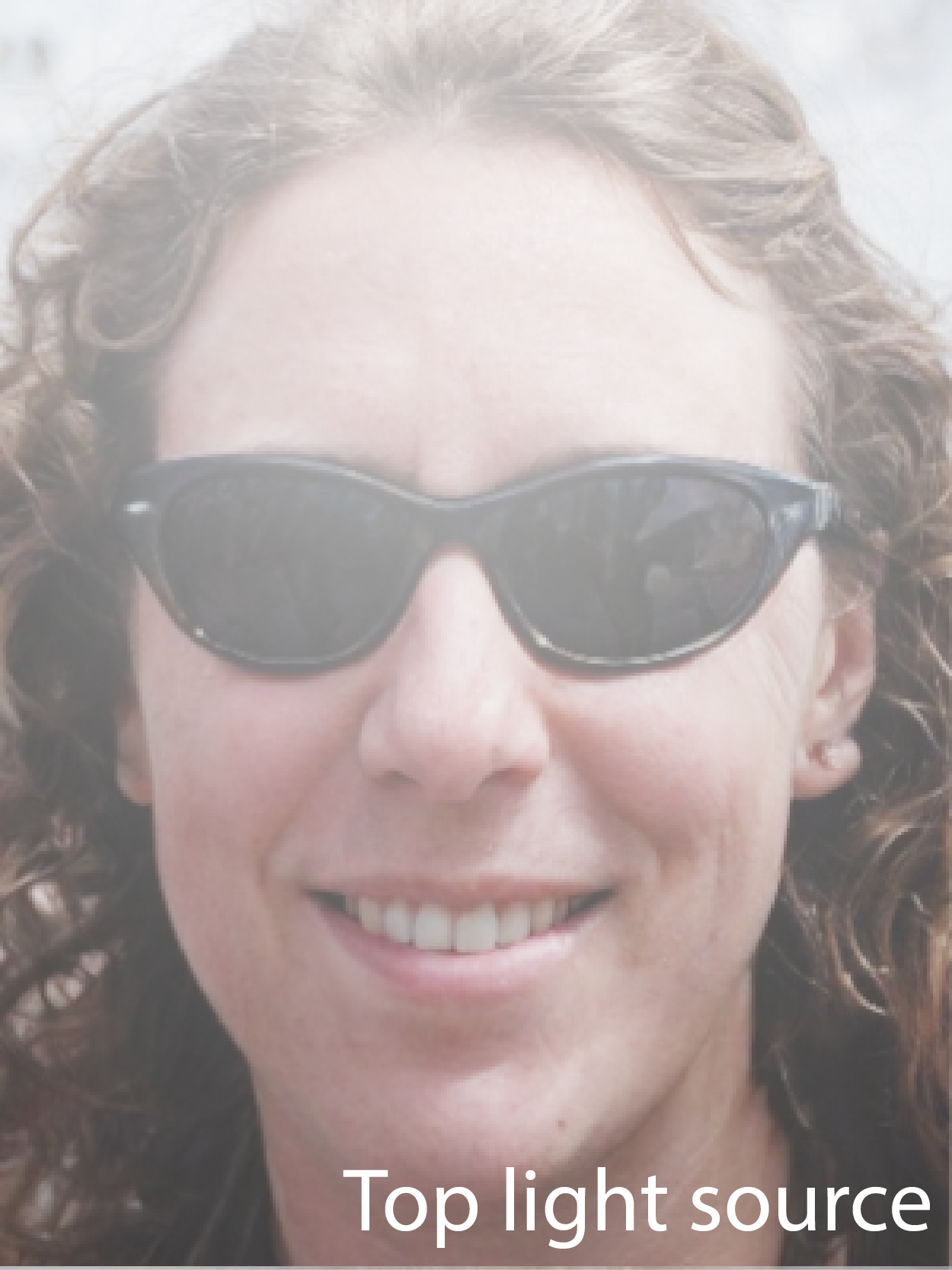}
  \end{minipage}
  \hfill
  \begin{minipage}[b]{0.116\textwidth}
    \includegraphics[width=\textwidth]{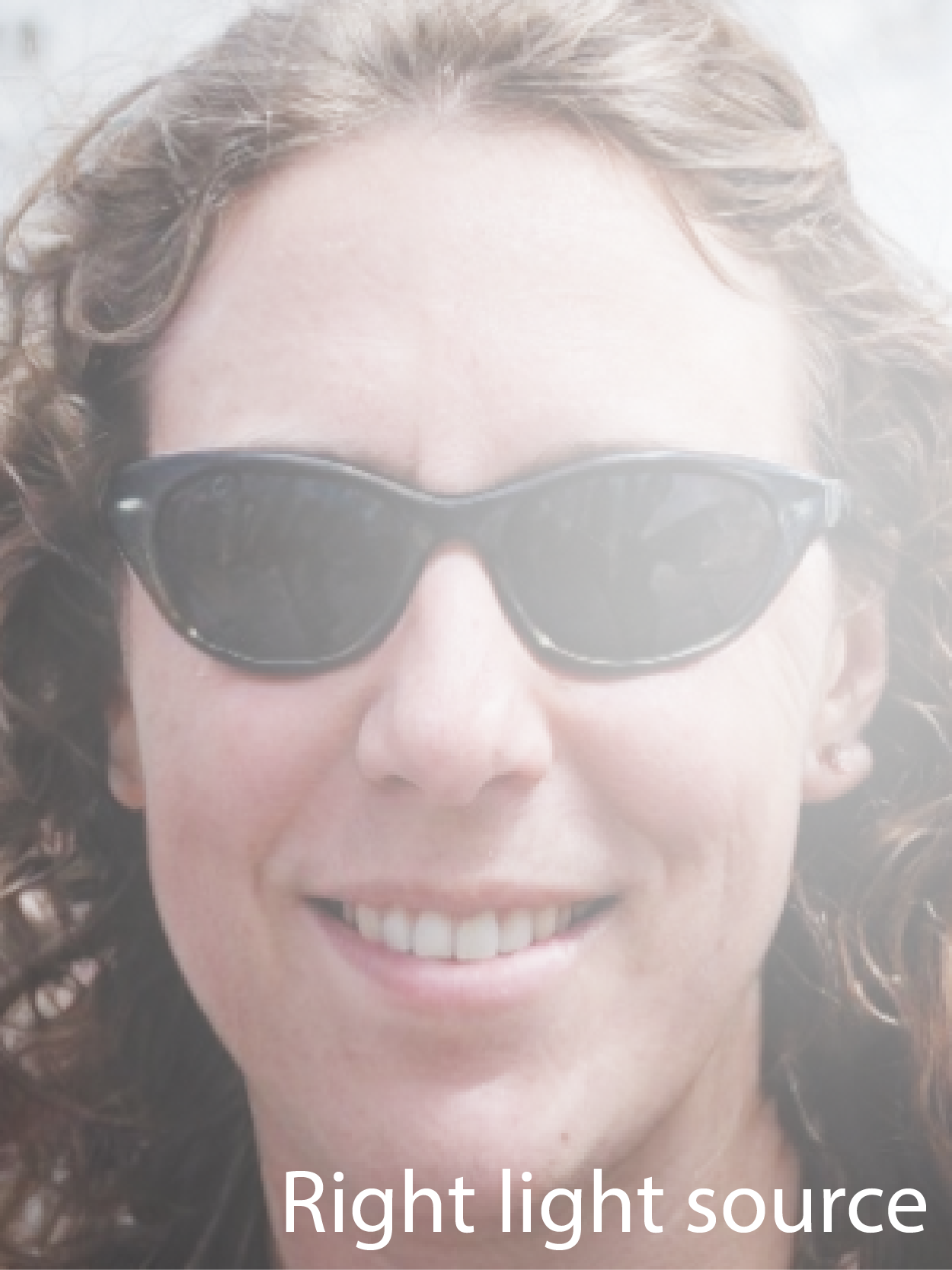}
  \end{minipage}
  \caption{Examples of samples from Experiment 3. From left to right: a face image, adding light source from left, adding light source from top, adding light source from right.} 
  \label{fig:p3_light}
  \vspace{-2mm}
\end{figure}

\begin{figure}[b]
  \centering
    \hfill
  \begin{minipage}[b]{0.116\textwidth}
    \includegraphics[width=\textwidth]{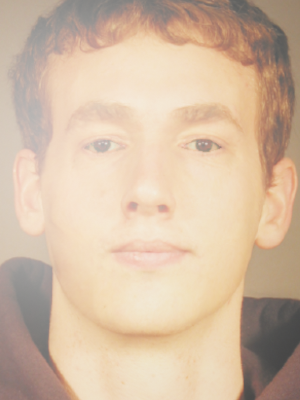}
  \end{minipage}
  \begin{minipage}[b]{0.116\textwidth}
    \includegraphics[width=\textwidth]{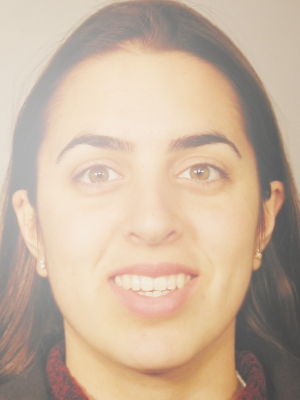}
  \end{minipage}
  \hfill
  \begin{minipage}[b]{0.116\textwidth}
    \includegraphics[width=\textwidth]{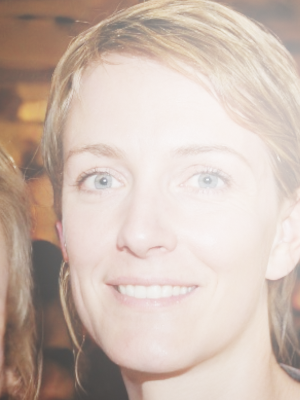}
  \end{minipage}
  \hfill
  \begin{minipage}[b]{0.116\textwidth}
    \includegraphics[width=\textwidth]{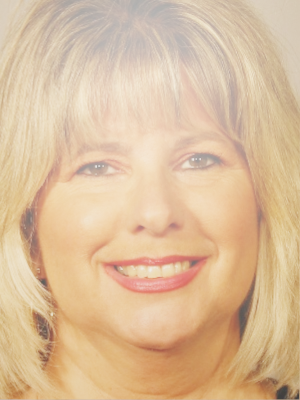}
  \end{minipage}
  \caption{Two sample pairs from Experiment 3. From left to right: synthetic face in Pair 1, generated by SREFI; real face in Pair 1; synthetic face in Pair 2, generated by StyleGAN; real face in Pair 2.}
  \label{fig:p3_sample}
\end{figure}
\begin{figure}[h]
  \vspace{-2mm}
  \centering
  \includegraphics[width=0.48\textwidth]{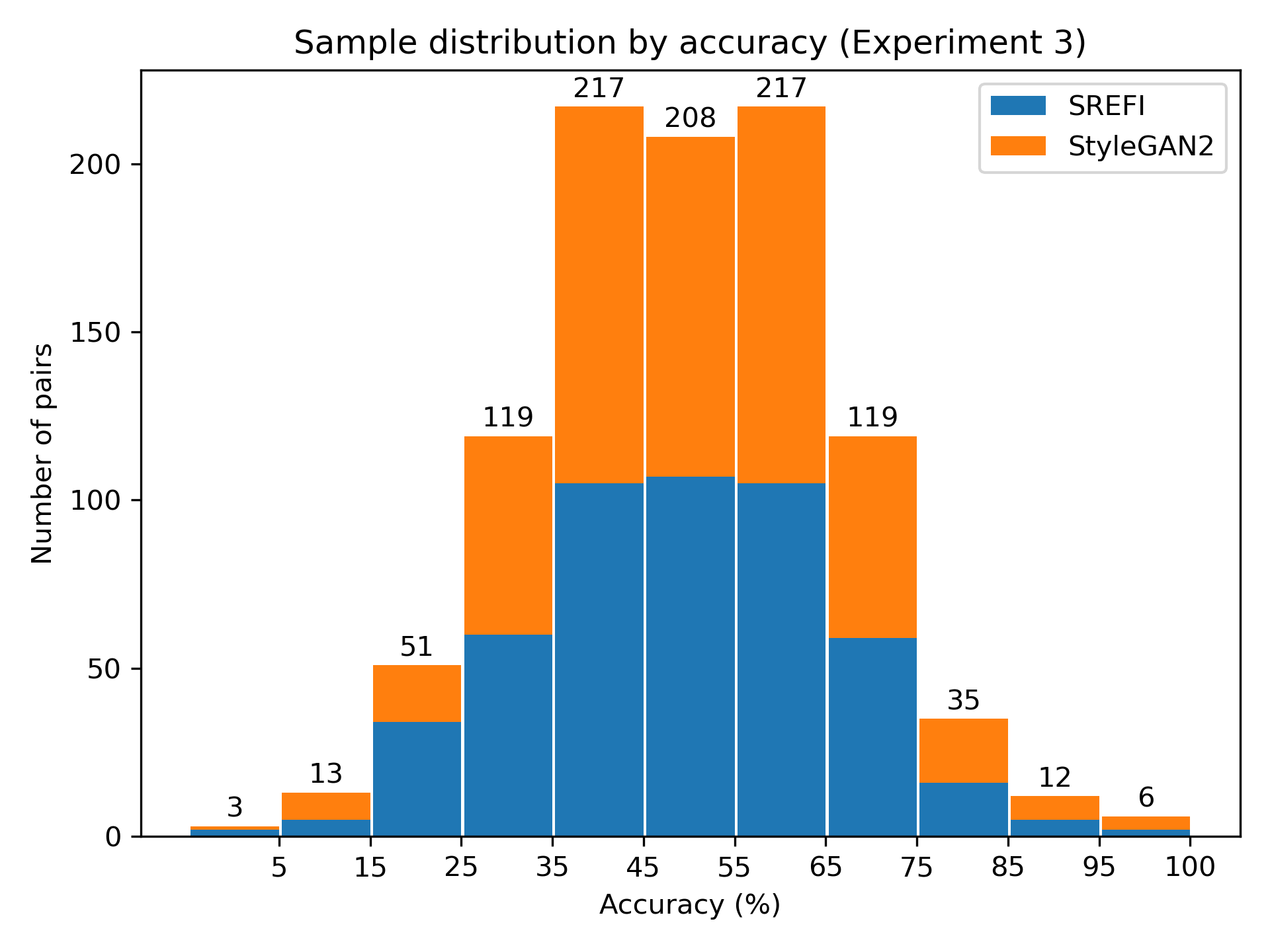}
  \caption{Human performance on 1,000 pairs of real / synthetic face images with lighting perturbation.}
  \label{fig:exp03_result}
  \vspace{-5mm}
\end{figure}

\subsubsection{Distinguishing faces under varied lighting}
We then moved on to examining whether light disrupts people's judgement. In this experiment, we evaluated the human ability to distinguish between real and synthetic faces under a changing light source. 1,000 real/synthetic faces were selected and pre-processed in the same manner as the first experiment. We then added an additional light source for each face image using a 3D rendering engine~\cite{Blender} to create a realistic sunlight effect. The direction of light is randomly selected among ``left'', ``top'' and ``right'', as shown in Fig.~\ref{fig:p3_light}. Sample face pairs are shown in Fig.~\ref{fig:p3_sample}. The human raters are asked the same questions as in the previous two experiments and the face images stay on the screen until human raters answer the question.



In total, 10,350 real / synthetic image pairs were evaluated in this experiment, and we ensured at least 10 human raters on Amazon's Mechanical Turk service examined each pair as in previous experiments. There were 172 different subjects involved in the evaluation and on average 10.35 people examined each pair of images. 
We also calculated the accuracy for all image pairs and the distribution of samples is shown in Fig.~\ref{fig:exp03_result}. Similar to the previous experiments, the data appear to be normally distributed, suggesting people are prone to mistaking synthetic faces for real faces. The  one-tailed one sample t-test results (Table~\ref{table: exp_result}) has high values of $p$ for Experiment 3, leaving the null hypothesis un-rejected. 

This result, combined with those from the previous two experiments, gives us a good snapshot of the current state of face synthesis technology. It is likely that human observers cannot gather enough information to solve this task in general. This finding has serious implications for the deployment of synthetic faces in many applications. We can safely say that the technology has developed to the point of being able to fool people in a reliable manner.
 

\subsection{Experiment 4: Do synthetic faces give away any information about real identities?}
Different from the previous experiments, in this experiment we focused on the question of the extent to which synthetic faces leak identity information related to any real faces used by the generation algorithms. This is another aspect of plausibility --- if synthetic faces look too much like the real faces they are generated from, they may be recognized as being real when technically not. Most face synthesis algorithms generate faces based on one or more real input faces. 

For example, SREFI generates synthetic images based on a primary face and 7-10 ``donor'' faces proximal to it in the CNN feature space. Therefore the output image preserves a portion of the features from the primary face, which brings the risk of identity leakage. For StyleGAN, potential identity leakage is also  a concern. Given latent representations of multiple face images from training, one way of creating a synthetic face image using StyleGAN is by using the average of the latent representations as the new face's latent representation and generating an image from it.  The latent representation of a face image captures the style information from coarse (\textit{e.g.}, face shape, pose) to fine (\textit{e.g.}, background color). Averaging the latent representation carries with it the risk of creating a new latent vector that highly resembles one of its components in some dimensions, which could lead to facial features that reveal one of the original faces the new faces was generated from.

\subsubsection{Data preparation}
In this experiment, we prepared four types of image pairs described as follows.

\textbf{500 same-real identity image pairs} are selected from the dataset that generated RPool1 in Section~\ref{sec:data}. We randomly selected 500 identities, each with two different frontal images of a neutral or happy expression. \textbf{500 different-real identity image pairs} are generated by randomly selecting two identities from RPool1, with each identity  contributing one frontal image to the pair. To avoid obviously easy tasks, all the pairs are of the same gender. We then used the same-real identities dataset to generate \textbf{500 real-synthetic (SREFI) identity image pairs}. 500 different real identities are selected as the base faces. For each real identity, 7 to 10 identities from the rest of the dataset are selected as donor faces to synthesize a new identity's face image~\cite{banerjee2017srefi}. 
Then the base face and the synthesized face constitute one real-synthetic pair. 

\begin{figure}[h]
  \centering
    \hfill
  \begin{minipage}[b]{0.47\textwidth}
    \includegraphics[width=\textwidth]{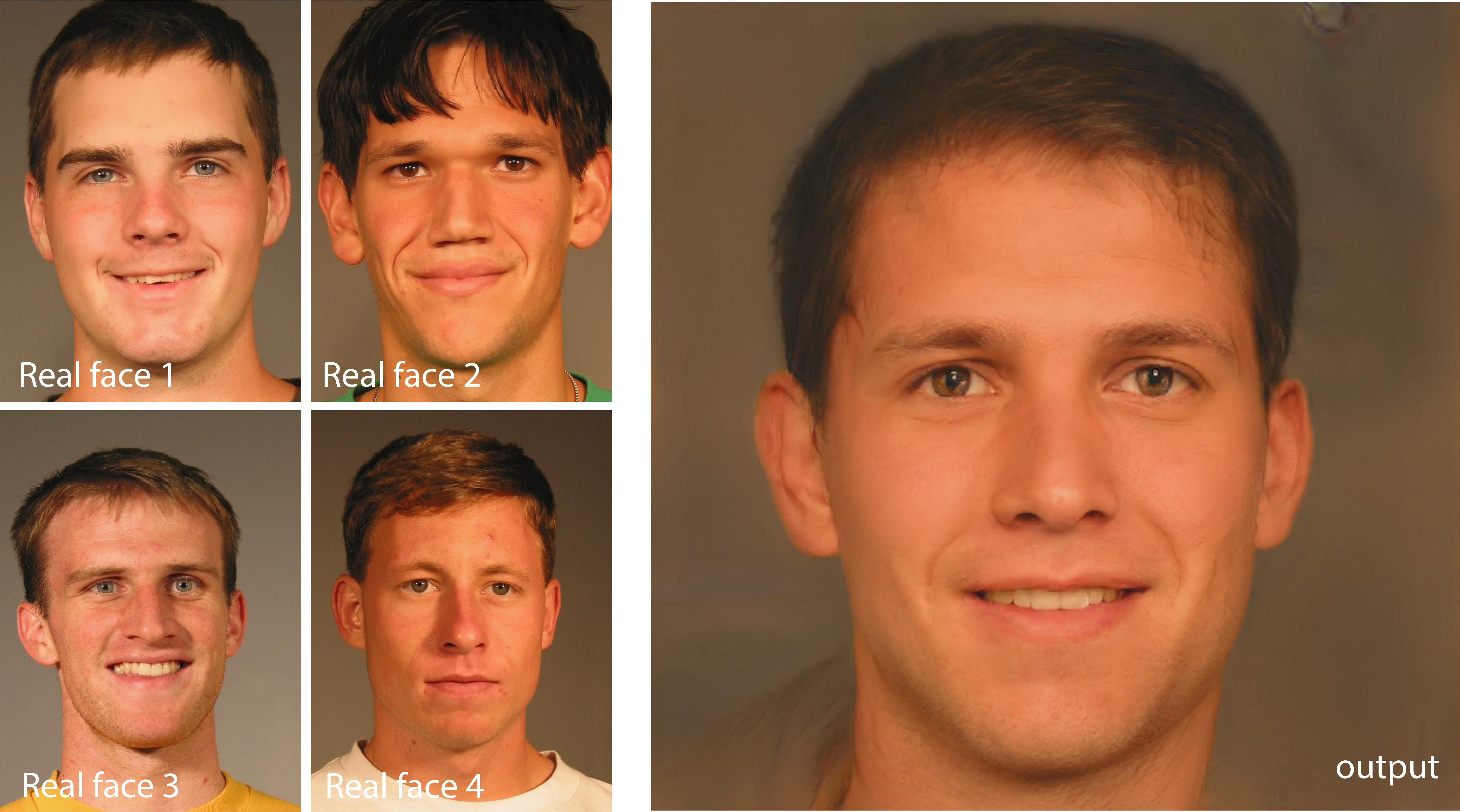}
  \end{minipage}
  \caption{Face image of a synthetic identity generated by StyleGAN. The output image on the right is generated by StyleGAN using the weighted average of the latent representations of the four real faces on the left.}
  \label{fig: stylegan_gen}
    \vspace{-3mm}
\end{figure}

\begin{figure}[h]
  \centering
\begin{subfigure}{0.11\textwidth}
  \includegraphics[width=\linewidth]{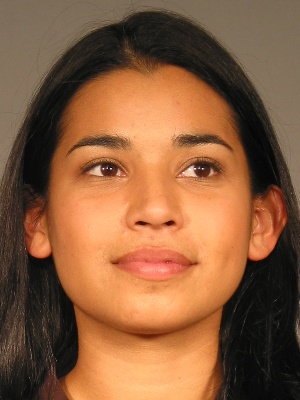}
  \caption{P1 real face}
  \label{fig:1}
\end{subfigure}\hfil 
\begin{subfigure}{0.11\textwidth}
  \includegraphics[width=\linewidth]{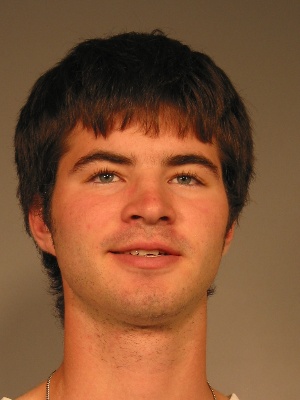}
  \caption{P2 real face}
  \label{fig:1}
\end{subfigure}\hfil 
\begin{subfigure}{0.11\textwidth}
  \includegraphics[width=\linewidth]{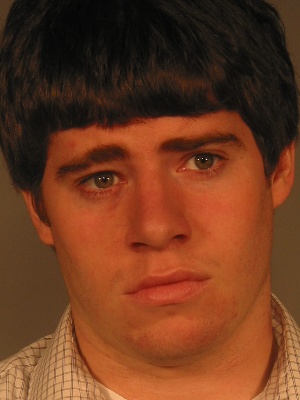}
  \caption{P3 real face}
  \label{fig:2}
\end{subfigure}\hfil 
\begin{subfigure}{0.11\textwidth}
  \includegraphics[width=\linewidth]{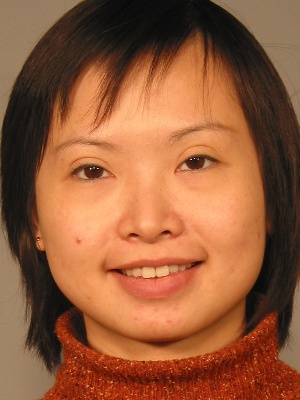}
  \caption{P4 real face}
  \label{fig:3}
\end{subfigure}

\medskip
\begin{subfigure}{0.11\textwidth}
  \includegraphics[width=\linewidth]{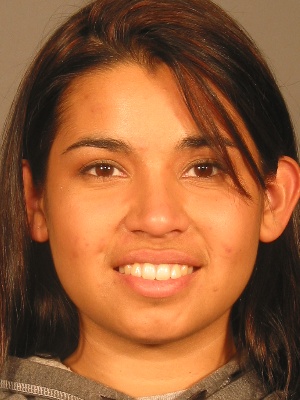}
  \caption{P1 real face from the same identity}
  \label{fig:4}
\end{subfigure}\hfil 
\begin{subfigure}{0.11\textwidth}
  \includegraphics[width=\linewidth]{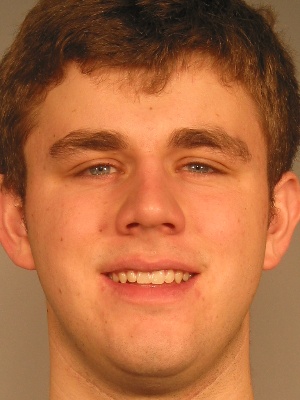}
  \caption{P2 real face from a different identity}
  \label{fig:4}
\end{subfigure}\hfil 
\begin{subfigure}{0.11\textwidth}
  \includegraphics[width=\linewidth]{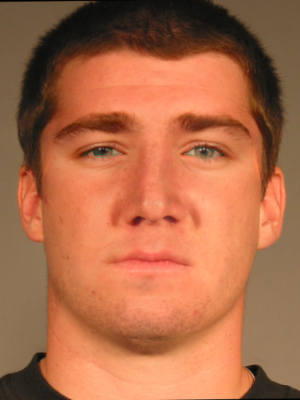}
  \caption{P3 face generatd with SREFI}
  \label{fig:5}
\end{subfigure}\hfil 
\begin{subfigure}{0.11\textwidth}
  \includegraphics[width=\linewidth]{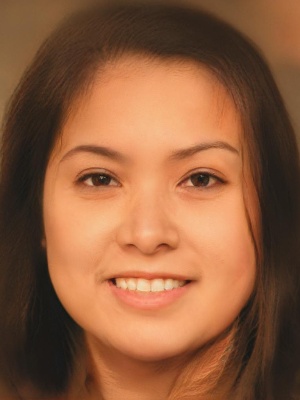}
  \caption{P4 face generated with StyleGAN}
  \label{fig: 6}
\end{subfigure}

  \caption{Four sample pairs from Experiment 4. Top row: real face images in Pairs 1-4. Bottom row (from left to right): another real face image from the same identity as the face in (a), real face image from an identity different from (b), synthetic face generated by SREFI using the real face (c) as an input face, synthetic face generated by StyleGAN with the real face (d) as one of the input faces.}
  \label{fig:p4_sample}
  \vspace{-5mm}
\end{figure}

To create the \textbf{500 real-synthetic (StyleGAN) identity image pairs}, we first randomly selected 500 real face images from RPool1 and calculated the corresponding latent representations using a pre-trained StyleGAN generator on the FFHQ dataset~\cite{karras2019style}. For each real face image, $N$-1 other real faces are randomly selected from the rest of the 500 images set. Following Equation~\ref{equation_latent}, the latent representation of the synthetic face to be created is the weighted combination of the latent representations $L_i$ of the $N$ real faces. Then the pre-trained StyleGAN generator is able to generate a face image from noise based on the latent representation. An example of a StyleGAN synthesized face is shown in Fig.~\ref{fig: stylegan_gen}. We set the variable $N$ to 4, and the coefficient $\alpha$ is randomly selected to be between 0.2 and 0.3. Thus each real face contributes equally to the output synthetic face while preserving certain face features. One of the base faces and the synthesized face constitute one real-synthetic pair.
\begin{equation}
     L_{output} =  \sum_{i=1}^{N} \alpha \times {L_{i}}
     \label{equation_latent}
\end{equation}
Fig.~\ref{fig:p4_sample} shows samples of the four types of image pairs.

\subsubsection{Experiment}
\begin{figure}[h]
  \centering
    \hfill
  \begin{minipage}[b]{0.48\textwidth}
    \includegraphics[width=\textwidth]{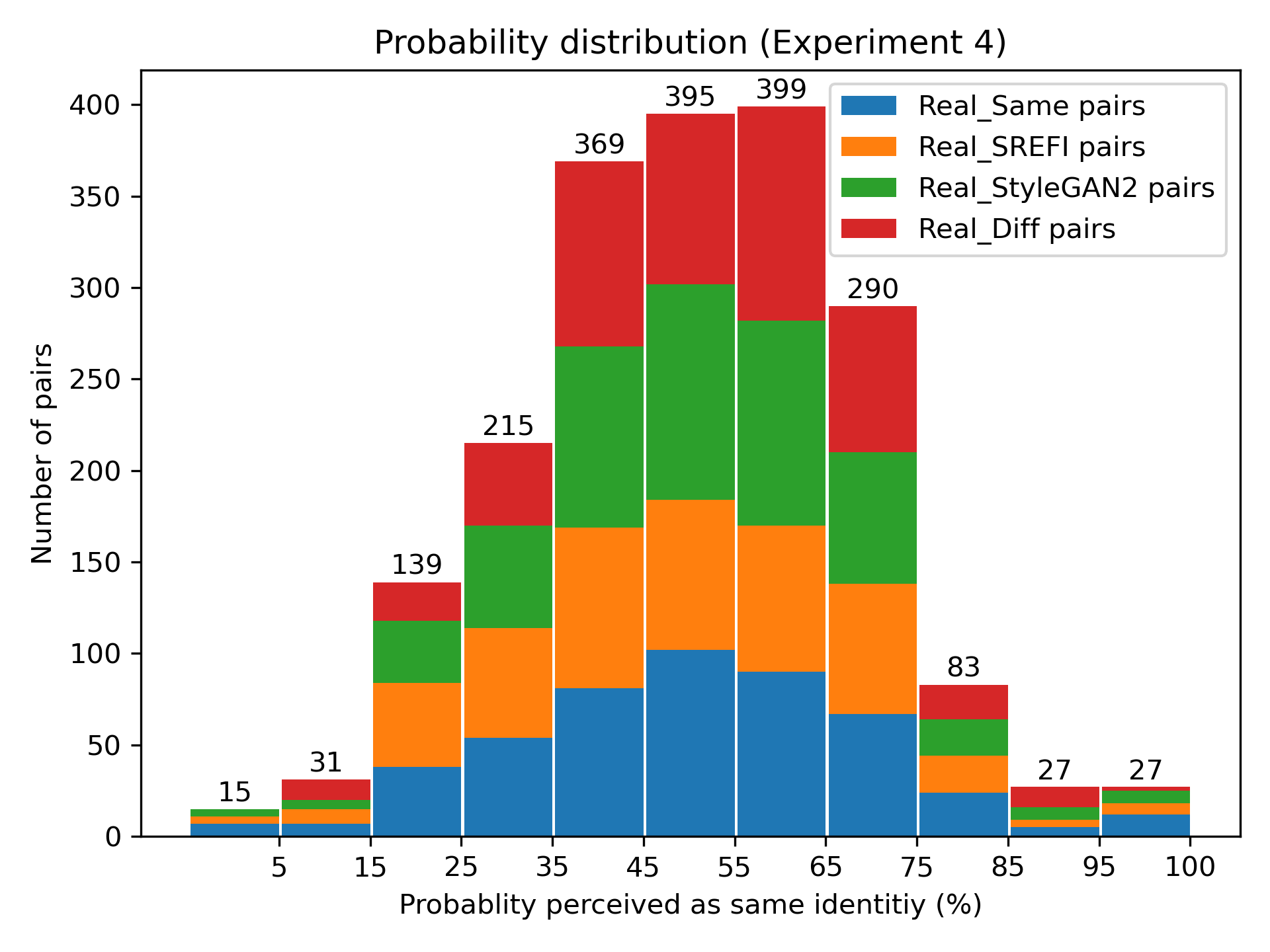}
  \end{minipage}
  \caption{Results of Experiment 4. Human performance on 500 same-real identities, 500 different-real identities, 500 real-SREFI identities, and 500 real-StyleGAN identities.}
  \label{fig: exp_04}
    \vspace{-6.5mm}
\end{figure}

Participants are presented with a pair of images and prompted with a question asking ``Are these faces of the same person or are they of different people?'' They are forced to choose between ``Same'' and ``Different''. Each participant sees a block of 150 image pairs (50 images for each type of pairs), and a total of 66 workers on Mechanical Turk participated in this task. Each pair of images was evaluated by 8.53 people on average. The distribution of pairs with respect to the probability of being considered from the same identity is shown in Fig.~\ref{fig: exp_04}.

For the real faces from the same identities, participants have a probability of 0.503 ($\pm 0.191$) of determining if the two images are from the same person. For the real faces of different identities, the probability of them being identified as the same person is 0.522 ($\pm 0.166$). The same probability value for the real-synthetic SREFI pairs is 0.491($\pm 0.185$) and 0.507($\pm 0.173$) for the real-synthetic StyleGAN pairs. These results are surprising because there is no discernible difference between the real image pairs and the real-synthetic (SREFI or STYLEGAN) pairs.
However, referring back to the literature on facial perception, it is well known that that unfamiliar face recognition is significantly more difficult than familiar face recognition --- even more so when no training period is given~\cite{johnston2009familiar}. This is a plausible explanation, which bolsters the case for the effectiveness of synthetic faces. 


%% file: 6conclusions.tex
\section{DISCUSSION}

The first three human perception experiments that were conducted in this study showed that the human ability to accurately distinguish between synthetic and real faces is not better than chance. What implication does this have for social media and other applications on the Internet, and what can we do to mitigate the negative impact of synthetic faces? This type of content has already appeared on social networks like LinkedIn~\cite{Satter_2019} in attempts to develop relationships with users who may possess sensitive information, while evading detection through reverse image searches. The next logical place for them to appear is in romance scams conducted via dating apps, where low-level criminal activity is already present~\cite{Justice_2021}. If a scammer is using high quality synthetic faces, our results indicate that it is probable that the average user will not notice that the persona they are interacting with isn't real based on the profile photo. All is not lost, however. Proposals to add fingerprints to synthetically generated content via the tools used to create it exist~\cite{yu2020artificial}, and effective detectors are being developed by the media forensics community~\cite{verdoliva2020media}. The latter is likely our best hope in assisting people in identifying such content in the wild.
